\documentclass{article}

\PassOptionsToPackage{table}{xcolor}
\usepackage{report}

\usepackage[utf8]{inputenc} % allow utf-8 input
\usepackage[T1]{fontenc}    % use 8-bit T1 fonts
\usepackage{hyperref}       % hyperlinks
\usepackage{url}            % simple URL typesetting
\usepackage{booktabs}       % professional-quality tables
\usepackage{amsfonts}       % blackboard math symbols
\usepackage{nicefrac}       % compact symbols for 1/2, etc.
\usepackage{microtype}      % microtypography
\usepackage{xcolor}         % colors

\usepackage{latexsym}
\usepackage{graphicx}
\usepackage{amsmath}
\usepackage{amssymb}
\usepackage{booktabs}
\usepackage{multirow}
\usepackage{array}
\usepackage{tikz}
\usepackage{pgfplots}
\pgfplotsset{compat=1.18}
\usepgfplotslibrary{fillbetween}
\usetikzlibrary{positioning,arrows.meta,calc,shapes.geometric,patterns,fit,backgrounds,decorations.pathreplacing,shapes}
\usepackage{float}
\usepackage[font=small]{caption}
\usepackage{capt-of}
\usepackage{hyperref}
\usepackage{geometry}
\usepackage{natbib}
\usepackage{enumitem}
\usepackage{xcolor}
\usepackage{dblfloatfix}
\usepackage{placeins}
\usepackage{algorithm}
\usepackage{algorithmic}
\usepackage{wrapfig} 
\usepackage{tcolorbox}
\tcbuselibrary{breakable}

\definecolor{cBlue}{HTML}{3B82F6}
\definecolor{cGreen}{HTML}{10B981}
\definecolor{cOrange}{HTML}{F59E0B}
\definecolor{cRed}{HTML}{EF4444}
\definecolor{cPurple}{HTML}{8B5CF6}
\definecolor{cGray}{HTML}{6B7280}
\definecolor{cDark}{HTML}{1F2937}
\definecolor{cLight}{HTML}{F3F4F6}
\definecolor{njuPurple}{RGB}{220,205,230}
\definecolor{njuPurpleLight}{RGB}{250,245,252}

\newtcolorbox{abstractbox}{
    colback=njuPurpleLight,
    colframe=njuPurple,
    boxrule=1pt,
    arc=4mm,
    left=8pt,
    right=8pt,
    top=8pt,
    bottom=8pt,
    opacityback=0.95,
    breakable
}

\usepackage{tikz}
\usepackage{pgfplots}
\usepgfplotslibrary{groupplots}
\pgfplotsset{compat=1.18}

\usepackage{etoolbox}

\usepackage{tabularx}
\let\oldmathbb\mathbb

\renewcommand{\mathbb}[1]{%
  \ifstrequal{#1}{1}{\mathbf{1}}{\oldmathbb{#1}}%
}
\title{Training Needs Trustworthy Worlds: Verified Synthetic Web Environments for Agent Learning}

\author{%
\begin{tabular}{ccc}
Chenghao Zhang$^{1}$ & Yuxi Cheng$^{2}$ & Saisai Hu$^{3}$
\end{tabular}\\[3pt]
\begin{tabular}{ccc}
Canran Xiao$^{4}$ & Yikai Mao$^{1}$ & Dan Roth$^{1}$
\end{tabular}
}

\begin{document}

\maketitle

\let\oldthefootnote\thefootnote
\let\thefootnote\relax
\footnotetext{$^{1}$~University of Pennsylvania.}
\footnotetext{$^{2}$~Unaffiliated.}
\footnotetext{$^{3}$~Pace University.}
\footnotetext{$^{4}$~Shenzhen Campus of Sun Yat-sen University.}
\let\thefootnote\oldthefootnote

\begin{abstract}
\begin{abstractbox}
Web agents promise to automate complex digital workflows, but their training remains limited by synthetic environments that look plausible while hiding broken links, inconsistent states, or infeasible tasks. 
We address the gap between scalable environment generation and trustworthy agent learning by constructing synthetic web environments that are executable, auditable, and grounded in backend state. 
Our framework represents each generated website as a structured scaffold of pages, navigation links, database records, state-change markers, and task constraints, then verifies and repairs structural, semantic, consistency, and feasibility defects before policy training. 
During interaction, ordinary UI transitions are executed deterministically, while persistent backend updates are invoked only through validated state-change markers, enabling dense rewards compiled from verified task-progress predicates. 
Across 500 synthetic environments spanning six domains, our method reduces task-blocking defects and improves feasible-task rate from 48.6\% to 94.8\%, while producing stronger PPO policies and improving transfer to WebArena, WebShop, and MiniWoB++ without LLM calls at evaluation time. 
These results show that verified synthetic environments can serve as a scalable and reliable training substrate for compact web agents, shifting synthetic web-agent learning from surface-level plausibility toward executable, state-grounded supervision.
\end{abstractbox}
\end{abstract}

\section{Introduction}
\label{sec:introduction}

Autonomous web agents aim to complete user-specified tasks by perceiving graphical interfaces, selecting grounded actions, and manipulating persistent web states. This capability is increasingly important for automating digital workflows such as shopping, booking, form submission, customer-service operations, and enterprise knowledge work. Despite rapid progress in large language and multimodal agents, web interaction remains challenging because successful execution requires long-horizon planning, precise grounding in dynamic interfaces, and reliable reasoning over hidden backend states rather than only visible page content.

Recent benchmarks and systems have substantially advanced web-agent research. Early platforms such as World of Bits and MiniWoB++ established reproducible browser-control tasks for reinforcement learning \citep{pmlr-v70-shi17a,liu2018reinforcementlearningwebinterfaces}, while WebShop, Mind2Web, WebArena, VisualWebArena, WorkArena, and BrowserGym introduced more realistic tasks, websites, demonstrations, and evaluation protocols \citep{yao2023webshopscalablerealworldweb,deng2023mind2webgeneralistagentweb,zhou2024webarenarealisticwebenvironment,koh2024visualwebarena,drouin2024workarena,dechezelles2024browsergym}. In parallel, LLM-based agents such as ReAct, WebVoyager, AgentOccam, Agent Q, and WebRL have improved reasoning, exploration, and learning from interaction traces \citep{yao2023reactsynergizingreasoningacting,he2024webvoyager,yang2024agentoccam,putta2024agentq,qi2024webrl}. These works demonstrate that web agents can benefit from richer environments and larger-scale interaction data. However, they also expose a central bottleneck: high-quality web-agent training requires many executable, stateful, and verifiable environments, yet realistic web environments are expensive to build, difficult to reset, and often limited in task coverage.

A natural direction is to use LLMs to synthesize environments, tasks, or agent experiences. Prior work has shown that generated environments and synthetic trajectories can support agent learning in embodied and digital settings \citep{zala2024envgengeneratingadaptingenvironments,patel2024largelanguagemodelsselfimprove}. Nevertheless, directly generated web environments are prone to a deeper reliability problem: locally plausible pages may still contain globally invalid workflows. For example, links may be unreachable, database values may contradict rendered content, required controls may be missing, and task success conditions may not correspond to any executable state transition. Such defects are especially harmful for reinforcement learning, because the agent may receive supervision from environments whose apparent failures are caused by scaffold errors rather than policy mistakes. Existing approaches do not fully resolve this high-level mismatch between \emph{plausible generation} and \emph{executable, state-grounded interaction}.

\begin{figure}[H]
    \centering
    \includegraphics[width=0.72\linewidth]{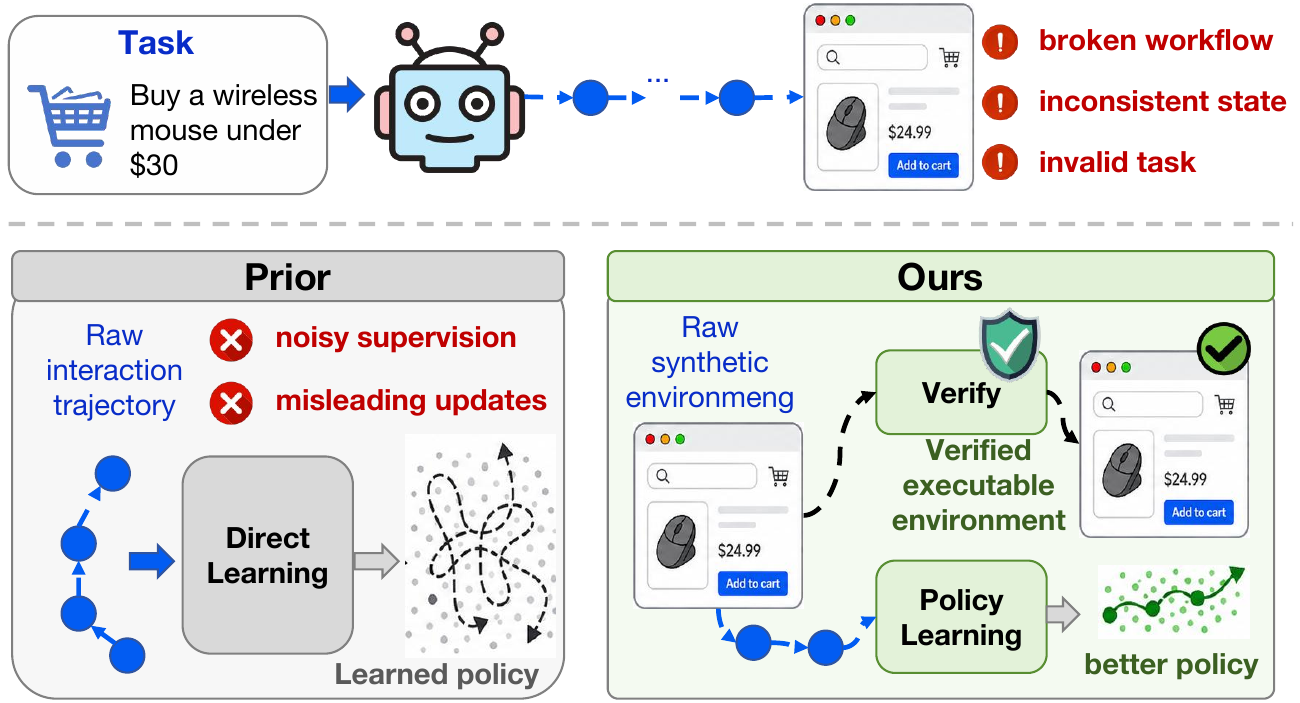}
    \caption{
        %\textbf{Verified synthetic web environments for reliable policy learning.}
        Prior methods train on superficially plausible but inconsistent web interactions, while our approach verifies and repairs the environment to provide executable, state-grounded supervision for more effective agent policies.
    }
    \label{fig:teaser}
\end{figure}

This paper asks: \emph{How can we construct synthetic web environments that are not only diverse and realistic, but also executable, auditable, and useful for training compact web agents?} We address this question by shifting the object of generation from isolated pages or trajectories to verified environment scaffolds. As shown in Fig.\ref{fig:teaser}, our approach first represents a synthetic website as a structured environment with pages, navigation, data records, state-changing events, and task constraints; then it verifies and repairs the scaffold before policy learning; finally, it trains agents using rewards derived from backend-grounded task progress rather than from surface-level textual judgments. This design aims to make synthetic web interaction a reliable training substrate, where failures are attributable to the policy instead of hidden environment invalidity.

Our contributions are summarized as follows:
\begin{itemize}
    \item We identify a core limitation of trajectory- or page-level generation: web-agent training depends on whether tasks are executable under consistent backend dynamics, not merely whether the generated interface appears plausible.

    \item We propose a framework that canonicalizes, verifies, and repairs generated web scaffolds, producing environments with explicit navigation structure, database bindings, state-change events, and task constraints.

    \item By deriving dense rewards from verified backend-state progress and executing ordinary interactions without LLM calls at evaluation time, our framework enables efficient reinforcement learning in synthetic web environments while preserving task-level auditability.
\end{itemize}

\section{Related Work}

\paragraph{Web interaction benchmarks and web agents.}
Early work framed web interaction as language-conditioned control over semi-structured interfaces, from instruction following to reproducible environments such as World of Bits and MiniWoB++ \citep{branavan-etal-2009-reinforcement,pmlr-v70-shi17a,liu2018reinforcementlearningwebinterfaces}. Later benchmarks expanded realism and task diversity, including WebShop, Mind2Web, WebArena, VisualWebArena, WorkArena, and BrowserGym \citep{yao2023webshopscalablerealworldweb,deng2023mind2webgeneralistagentweb,zhou2024webarenarealisticwebenvironment,koh2024visualwebarena,drouin2024workarena,dechezelles2024browsergym}. Recent agents such as WebVoyager and AgentOccam further show that improved observation and action design can benefit web navigation \citep{he2024webvoyager,yang2024agentoccam}. However, these efforts mainly provide fixed benchmarks or agent-side improvements, rather than methods for automatically constructing large-scale, executable, stateful, and verifiably valid training environments. Our work instead generates checked web environments whose pages, data bindings, events, and task constraints are explicitly represented and verified before policy learning.

\paragraph{Synthetic environments and agent experience generation.}
A complementary line of work uses LLMs to reduce the cost of interactive learning by generating environments, tasks, or trajectories. EnvGen uses LLMs to adaptively create training environments for embodied RL agents \citep{zala2024envgengeneratingadaptingenvironments}, while DreamGym synthesizes agent experiences through a learned reasoning-based environment model \citep{chen2025scalingagentlearningexperience}. For web agents, self-improvement methods synthesize or refine interaction data from agent rollouts, including model-generated web trajectories \citep{patel2024largelanguagemodelsselfimprove}, search- and preference-based refinement in Agent Q \citep{putta2024agentq}, and self-evolving curriculum RL in WebRL \citep{qi2024webrl}. These methods show the promise of using generated interaction data, but the generated supervision is often trajectory-centric and may inherit invalid dynamics, inconsistent states, or unverifiable task completions. Our method instead treats the environment itself as the object of generation: raw scaffolds are canonicalized, verified, repaired, and instrumented with constrained state-change markers, so that rollouts are produced by an executable simulator rather than by unconstrained trajectory synthesis.

\paragraph{Reward, verification, and policy learning for digital agents.}
Learning web agents is difficult because long-horizon tasks often provide sparse terminal feedback. Prior RL methods improved exploration with demonstrations and workflow constraints \citep{liu2018reinforcementlearningwebinterfaces}, while WebShop used programmatic matching functions to provide task-specific rewards \citep{yao2023webshopscalablerealworldweb}. More recent work studies automatic evaluators and process reward models for web or device-control agents, such as autonomous evaluator-guided refinement and Web-Shepherd's step-level reward modeling \citep{pan2024autonomousevaluation,chae2025webshepherd}. These approaches improve feedback quality but often rely on learned or LLM/VLM-based evaluators, which can be costly, miscalibrated, or disconnected from the true backend state. Our work takes a more environment-grounded route: task constraints are compiled into backend-state predicates that yield dense progress rewards, while invalid marker-triggered updates are rejected by construction. This enables standard PPO training \citep{schulman2017proximalpolicyoptimizationalgorithms} of compact policies in checked synthetic environments, without requiring LLM calls during policy evaluation.

\section{Preliminaries}
\label{sec:preliminaries}

\subsection{Web Interaction and Environment Scaffold}
\label{sec:preliminaries_web}

We model web interaction as a goal-conditioned sequential decision process. A task \(t\in T\) provides an instruction \(x_t\) and a programmatic completion constraint \(C_t\) evaluated on the environment state. Starting from \(s_0\sim\rho_t\), the agent observes a rendered interface, selects a DOM-grounded action, and triggers a state transition:
\begin{equation}
o_k=\mathrm{Render}(s_k),\qquad
a_k\in A(o_k),\qquad
s_{k+1}=\mathcal T(s_k,a_k).
\end{equation}
The terminal success signal is given by \(C_t(s)\in\{0,1\}\). This setting is difficult because the agent only observes the rendered interface rather than the full backend state, and successful completion often requires long-horizon navigation, form filling, and stateful updates.

We represent a generated web environment as
$
E=(P,L,D,T),
$
where \(P\) is the page set, \(L\subseteq P\times P\) is the navigation graph, \(D\) is the database schema with initialized records, and \(T\) is the task set with verifiable constraints. Each page \(p_i\in P\) contains a DOM tree \(\tau_i\) and interactive elements \(I_i\), whose action types and argument schemas ground high-level actions in executable UI operations.

This representation separates two layers. The \emph{static scaffold} defines pages, links, database bindings, interactive elements, and task constraints before execution. The \emph{dynamic layer} specifies how actions update persistent records and session variables during interaction. This separation is necessary because one-shot LLM generation often produces interfaces that are locally plausible but globally invalid. We consider four scaffold defects: structural defects \((\Delta_S)\), such as broken links or unreachable pages; semantic defects \((\Delta_C)\), such as invalid field values or placeholder content; consistency defects \((\Delta_X)\), such as contradictory entity attributes across pages; and feasibility defects \((\Delta_T)\), where missing controls or workflow steps make a task unsatisfiable.

\subsection{Motivating Diagnostics}
\label{sec:preliminaries_diagnostics}

We run a lightweight diagnostic study on 500 raw LLM-generated environments from six domains, covering 6,842 generated tasks, and analyze 1,800 successful trajectories from scripted and human-assisted rollouts. For defect category \(c\), let \(N_c(E)\) be its per-environment count. We measure its task-blocking effect as
\begin{equation}
B_c(E)=
\frac{1}{|T(E)|}
\sum_{t\in T(E)}
\mathbb{1}\!\left[\operatorname{Block}_c(E,t)=1\right],
\end{equation}
where \(\operatorname{Block}_c(E,t)\) indicates that task \(t\) is blocked by category \(c\) under bounded trace analysis. For a trajectory \(\tau=(s_0,a_0,\ldots,s_H)\), we measure marker-trigger sparsity by
\begin{equation}
M(\tau)=
\frac{1}{H}
\sum_{k=0}^{H-1}
\mathbb{1}\!\left[\mu(p_k,a_k)\neq \varnothing\right],
\qquad
U(\tau)=
\sum_{k=0}^{H-1}
\mathbb{1}\!\left[\mu(p_k,a_k)\neq \varnothing\right],
\end{equation}
where \(\mu(p_k,a_k)\) returns the state-change marker triggered by action \(a_k\).

\begin{figure*}[t]
    \centering
    \includegraphics[width=\textwidth]{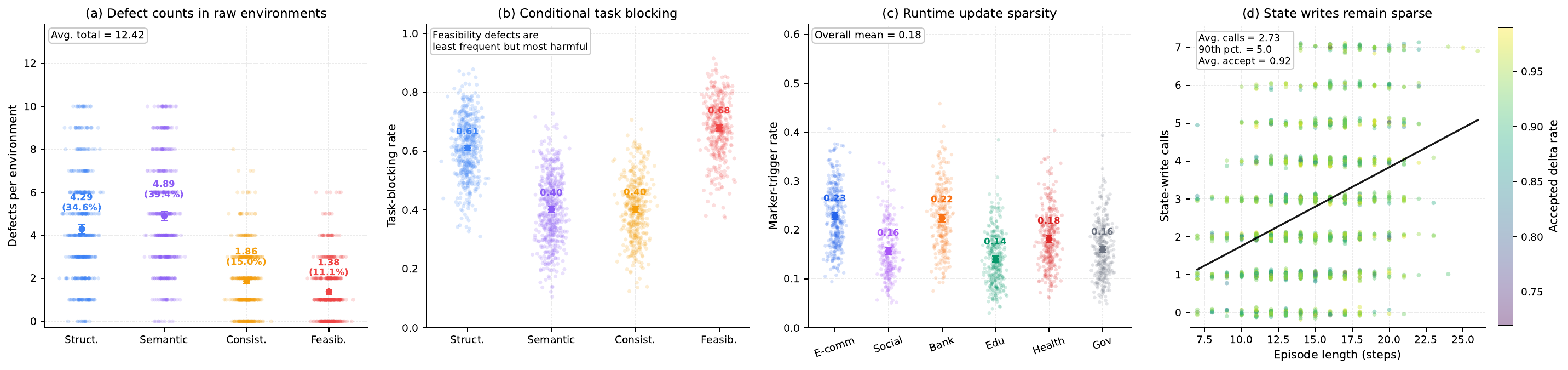}
    \caption{\textbf{Motivating diagnostics.}
    \textbf{(a)} Raw LLM-generated environments contain frequent defects, with semantic and structural defects dominating the average total of 12.4 defects per environment.
    \textbf{(b)} Defect frequency and harmfulness differ: feasibility defects are less frequent but most likely to block task completion.
    \textbf{(c)} Marker-triggered updates are sparse across domains, indicating that most interactions are deterministic UI transitions.
    \textbf{(d)} The number of state-write calls remains small even for longer episodes.}
    \label{fig:prelim_diagnostics}
\end{figure*}

Figure~\ref{fig:prelim_diagnostics} supports two design choices. First, raw scaffolds require verification because defect counts alone understate task-level harm: feasibility and structural defects are especially likely to block executable workflows. Second, runtime state updates are sparse: most steps are navigation, inspection, scrolling, or local input editing, while persistent backend writes occur only at a few marker-triggered events. These observations motivate verified static scaffolds and event-driven dynamic updates. Additional details on bounded trace analysis, defect attribution, marker statistics, and plotting protocol are provided in Appendix~\ref{app:prelim_diagnostic_details}.

\subsection{Dynamic State and Event-Driven Updates}
\label{sec:preliminaries_dynamic}

The dynamic layer maintains the database instance and session state used for rendering and access control. Ordinary actions, such as navigation, scrolling, menu expansion, and local text entry, are executed deterministically. Only marked events, such as \texttt{add-to-cart}, \texttt{submit-order}, or \texttt{update-profile}, invoke a constrained state writer to propose a backend delta. The delta is committed only if it satisfies marker preconditions, schema constraints, and environment invariants. Thus, the simulator avoids calling a generative model at every step while keeping persistent state changes explicit, auditable, and aligned with the verified scaffold.

\section{Method}
\label{sec:method}

\begin{figure}[htb]
	\centering
	\includegraphics[width=\linewidth]{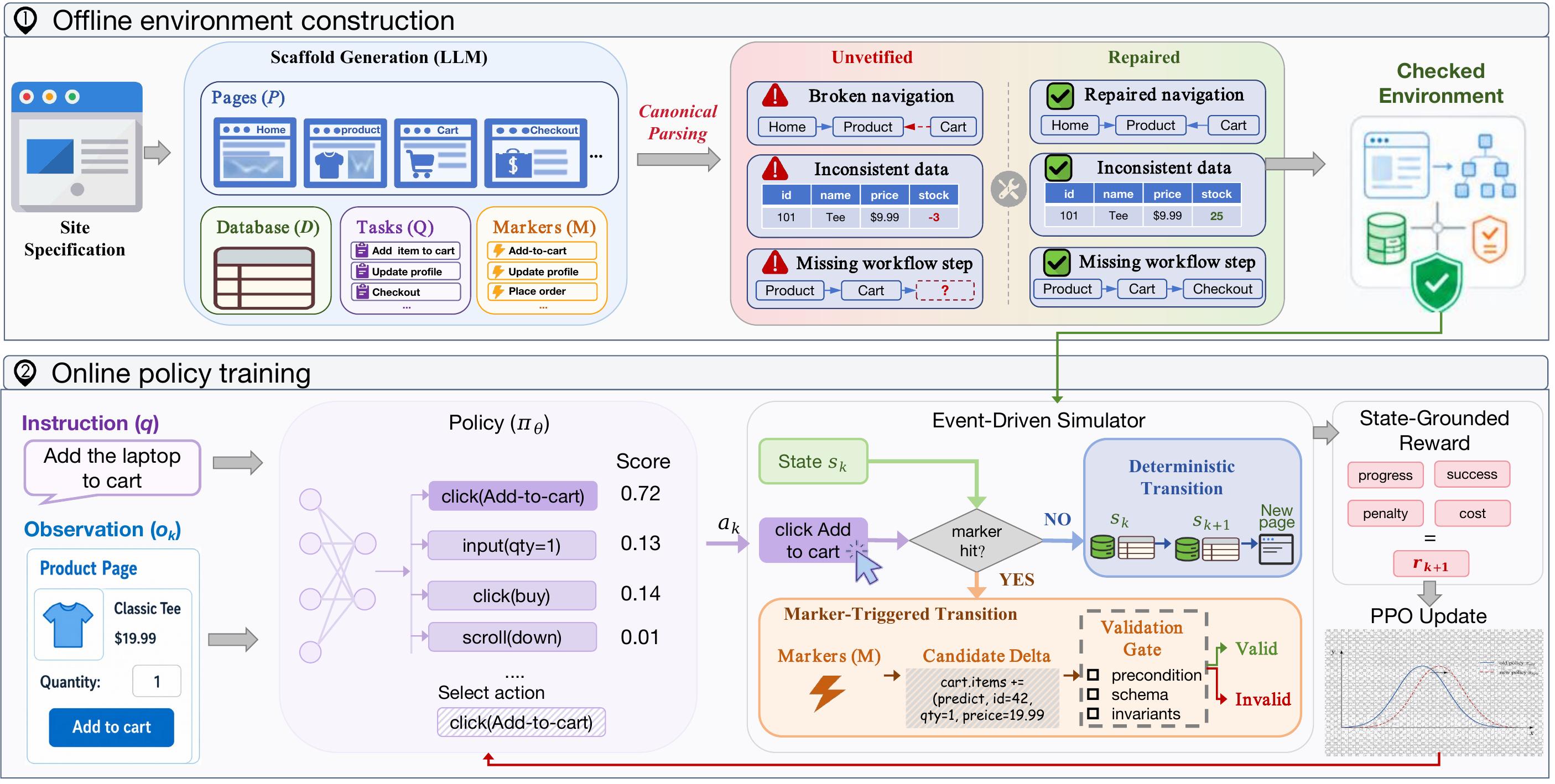}
	\caption{\textbf{Pipeline of verified synthetic web-environment construction and policy training.}
The offline stage generates a scaffold from a site specification, canonicalizes it into pages, data, tasks, and markers, and verifies/repairs navigation, data consistency, and workflow feasibility. The online stage trains a compact policy in the checked environment, where deterministic transitions and marker-validated state updates provide state-grounded rewards for PPO.}
	\label{fig:method_pipeline}
\end{figure}

Given a domain-level website description, our method constructs an executable synthetic web environment for training compact web agents. The environment is first represented as a structured scaffold with pages, links, database records, state-change markers, and task constraints. We then verify and repair the scaffold before training, so that generated tasks correspond to reachable workflows instead of artifacts of one-shot generation. During interaction, the simulator executes ordinary UI transitions deterministically and invokes constrained state writes only when a verified marker is triggered. Task constraints are compiled into backend-grounded progress predicates, which provide dense rewards for PPO training while keeping the learned policy independent of LLM calls at evaluation time.
Fig.\ref{fig:method_pipeline} shows the pipeline of our method.

\subsection{Environment Representation and Marker Instrumentation}
\label{sec:env_representation}

A generated web environment must support three operations throughout training: rendering observations, executing actions, and checking task completion against ground-truth state. We therefore represent the environment as a structured object instead of a collection of independent HTML pages. This representation exposes the causal chain from backend records to rendered DOM nodes, from UI events to state updates, and from state changes to task progress.

We represent an environment as
\begin{equation}
\mathcal E=(\mathcal P,\mathcal L,\mathcal D,\mathcal M,\mathcal Q),
\label{eq:env_rep}
\end{equation}
where \(\mathcal P\) is the page set, \(\mathcal L\subseteq\mathcal P\times\mathcal P\) is the navigation graph, \(\mathcal D\) is the database schema with initialized records, \(\mathcal M\) is the set of state-change markers, and \(\mathcal Q\) is the task set. Each task \(t\in\mathcal Q\) contains an instruction \(x_t\), an initial-state distribution \(\rho_t\), and a completion constraint \(C_t:\mathcal S\rightarrow\{0,1\}\) evaluated on simulator state.

At step \(k\), the simulator state, rendered observation, and executable action satisfy
\begin{equation}
s_k=(p_k,b_k,\sigma_k,h_k),\qquad
o_k=\mathrm{Render}(p_k,b_k,\sigma_k),\qquad
a_k\in\mathcal A(o_k).
\label{eq:state_obs_action}
\end{equation}
Here, \(p_k\) is the current page, \(b_k\) is the database instance, \(\sigma_k\) is the session state, \(h_k\) is the action history, and \(\mathcal A(o_k)\) is the DOM-grounded candidate action set extracted from the current observation. The rendering function fills page templates using database bindings and session variables, making the observation a deterministic view of the underlying state.

State-changing operations are specified by markers:
\begin{equation}
m=(e_m,\mathrm{pre}_m,R_m,W_m,\Omega_m).
\label{eq:marker}
\end{equation}
In Eq.~\eqref{eq:marker}, \(e_m\) is the triggering UI element, \(\mathrm{pre}_m\) is a precondition over the current state and action, \(R_m\) and \(W_m\) are the permitted read and write fields, and \(\Omega_m\) is the operation signature. A marker therefore constrains which backend fields are accessed and modified when a UI event commits a persistent update. Additional parsing rules and marker schemas are given in Appendix~\ref{app:canonical_verification}.

\subsection{Verification-Guided Scaffold Construction}
\label{sec:verification_repair}

One-shot LLM generation often produces locally plausible pages with globally invalid workflows. Broken links, inconsistent entity attributes, invalid database bindings, and infeasible tasks create misleading training signals. The verification stage detects these failures before policy learning; the repair stage modifies only the scaffold components implicated by accepted defect reports.

Given a site description \(q\), the generator produces raw scaffold artifacts \(\widetilde{\mathcal E}\), including page templates, navigation links, database records, task instructions, and completion constraints. A canonical parser converts \(\widetilde{\mathcal E}\) into \(\mathcal E^{(0)}\), extracts DOM elements and database bindings, and builds the initial marker set. Deterministic checks handle reachability, link integrity, schema validity, DOM--database binding, and marker read--write consistency. These checks are specified in Appendix~\ref{app:canonical_verification}.

At repair iteration \(r\), symbolic checks and semantic verifiers produce a defect set
\begin{equation}
\Delta^{(r)}
=
\Delta_{\mathrm{sym}}^{(r)}
\cup
\Delta_{\mathrm{str}}^{(r)}
\cup
\Delta_{\mathrm{sem}}^{(r)}
\cup
\Delta_{\mathrm{cons}}^{(r)}
\cup
\Delta_{\mathrm{feas}}^{(r)} .
\label{eq:defect_union}
\end{equation}
The five terms denote deterministic symbolic defects, structural defects, semantic content defects, cross-page consistency defects, and task feasibility defects. Each reported defect \(d\) is associated with a location, affected object, severity score, evidence, and verifier confidence. Reports with the same canonical key are merged. The aggregated confidence is
\begin{equation}
\mathrm{conf}(d)
=
\frac{
\sum_j \omega_j c_j(d)\mathbf{1}[d\in\Delta_j^{(r)}]
}{
\sum_j \omega_j \mathbf{1}[d\in\Delta_j^{(r)}]
},
\label{eq:defect_conf}
\end{equation}
where \(j\) indexes verifiers, \(c_j(d)\) is verifier \(j\)'s confidence for defect \(d\), \(\omega_j\) is the reliability weight of verifier \(j\), and \(\mathbf{1}[\cdot]\) is the indicator function. Defect-triggered verifier coordination is used to re-check only defect categories that affect one another; the coordination protocol is described in Appendix~\ref{app:verifier_coordination}.

Accepted defects are repaired in dependency order. We rank each defect by
\begin{equation}
\mathrm{score}(d)
=
\lambda_1\mathrm{sev}(d)
+
\lambda_2\log(1+\mathrm{dep}(d))
+
\lambda_3\mathrm{scope}(d)
-
\lambda_4\mathrm{cost}(d),
\label{eq:repair_score}
\end{equation}
where \(\mathrm{sev}(d)\) is severity, \(\mathrm{dep}(d)\) is the number of downstream defects that depend on \(d\), \(\mathrm{scope}(d)\) counts affected pages and tasks, \(\mathrm{cost}(d)\) estimates repair complexity, and \(\lambda_1,\ldots,\lambda_4\ge0\) are fixed scheduling weights. The repair operator updates the environment as \(\mathcal E^{(r+1)}=\mathcal R_{d^\star}(\mathcal E^{(r)})\), where \(d^\star\) is the highest-scoring defect whose dependencies have been satisfied. Structural repair modifies links and required elements; semantic repair rewrites content while preserving database bindings; consistency repair propagates canonical database values; feasibility repair inserts missing workflow steps and marker signatures. The loop terminates when no accepted critical defect remains and each task has a bounded executable trace satisfying its completion constraint. Repair operators and termination checks are given in Appendix~\ref{app:repair_ops}.

\subsection{Event-Driven Simulation}
\label{sec:event_simulation}

A training simulator must be efficient and inspectable. Most web interactions, such as navigation, scrolling, local text entry, and client-side validation, follow deterministic rules. Persistent changes, such as form submission, entity creation, profile update, and permission change, require backend writes. We therefore execute deterministic transitions by default and restrict generative state updates to verified marker-triggered operations.

Let \(\mu(p_k,a_k)\) return the marker triggered by action \(a_k\) on page \(p_k\), and let \(\varnothing\) denote no marker. The simulator transition is
\begin{equation}
(s_{k+1},\epsilon_k)
=
\begin{cases}
(F_{\mathrm{det}}(s_k,a_k),0),
& \mu(p_k,a_k)=\varnothing,\\
(F_{\mathrm{mark}}(s_k,a_k,\delta_k),\mathbf{1}[\delta_k=\bot]),
& m_k=\mu(p_k,a_k)\neq\varnothing .
\end{cases}
\label{eq:event_transition}
\end{equation}
Here, \(F_{\mathrm{det}}\) applies deterministic UI transitions, \(F_{\mathrm{mark}}\) applies a validated backend update, \(\delta_k\) is the state delta, and \(\epsilon_k\) indicates a rejected transition.

For a marker-triggered operation, a constrained state writer proposes a candidate delta using the current page, relevant database fragments, session state, action arguments, and marker signature. The candidate is accepted only after validation:
\begin{equation}
\delta_k
=
\mathrm{Validate}_{m_k}(\hat{\delta}_k;s_k,a_k)
\in
\mathcal U_{m_k}(s_k,a_k)\cup\{\bot\}.
\label{eq:delta_validate}
\end{equation}
The feasible update set \(\mathcal U_{m_k}(s_k,a_k)\) contains deltas satisfying the marker precondition, permitted read--write fields, database schema, and environment invariants. If validation returns \(\bot\), the simulator keeps the previous state and records a violation. This design makes every accepted persistent update traceable to a marker, a write set, and a validated state delta. The state writer interface and validation rules are detailed in Appendix~\ref{app:simulator_details}.

\subsection{State-Grounded Reward and Policy Learning}
\label{sec:reward_policy}

Terminal success alone gives sparse supervision for long-horizon web tasks. Since the checked environment exposes backend records and session variables, intermediate progress is computed from task constraints rather than from natural-language self-assessment. The reward therefore measures verified state progress and penalizes rejected transitions.
For each task \(t\), we compile its completion constraint \(C_t\) into progress predicates \(\Phi_t=\{\phi_{t,1},\ldots,\phi_{t,M_t}\}\), where each \(\phi_{t,i}:\mathcal S\rightarrow\{0,1\}\) checks a necessary intermediate condition. These conditions include visiting required pages, satisfying form constraints, creating target entities, updating correct attributes, and matching rendered views with backend state. The predicates define a progress potential:
\begin{equation}
\Psi_t(s)
=
\frac{1}{\sum_{i=1}^{M_t} w_i}
\sum_{i=1}^{M_t} w_i\phi_{t,i}(s),
\label{eq:progress_potential}
\end{equation}
where \(w_i\ge0\) is the importance weight of predicate \(\phi_{t,i}\). The policy observes \(o_k\) and \(x_t\), while \(C_t\), \(\Phi_t\), and backend state are used only by the environment to compute rewards.

The step reward is
\begin{equation}
r_k
=
\mathbf{1}[C_t(s_{k+1})=1]
+
\alpha\bigl(\Psi_t(s_{k+1})-\Psi_t(s_k)\bigr)
-
\gamma\epsilon_k
-
\eta .
\label{eq:reward}
\end{equation}
The first term gives terminal success reward, the second term gives net verified progress, the third term penalizes rejected state transitions, and \(\eta\) is a step cost. The constants \(\alpha,\gamma,\eta\ge0\) are fixed.

We train a compact policy \(\pi_\theta(a_k\mid o_k,x_t)\) using standard clipped PPO over DOM-grounded candidate actions. During synthetic training rollouts, marker-triggered simulator updates may use the constrained state writer in Eq.~\eqref{eq:delta_validate}; the learned policy itself performs action selection without LLM calls during evaluation. More details are provided in Appendix~\ref{app:ppo_details}.

\section{Experiments}
\label{sec:experiments}

We design experiments to answer three questions. 
First, does verification convert superficially plausible generated websites into executable training environments? 
Second, does training on verified environments improve compact policies beyond training on raw synthetic environments? 
Third, which components of the verification--repair--reward pipeline are responsible for the gains? We also discussed other questions regarding the mechanism of our method; see \S\ref{app_exp}.

\subsection{Experimental Setup}

\paragraph{Synthetic environment suite.}
We generate 500 synthetic web environments from domain-level site specifications across six domains: e-commerce, social media, banking, education, healthcare, and government. 
Each environment contains 15--30 pages, a navigation graph, database records, state-change markers, and a set of task instructions with programmatic completion constraints. 
Unless otherwise stated, we split environments by site specification into 350 training environments, 75 validation environments, and 75 held-out test environments, ensuring that held-out environments do not share page templates or task constraints with training environments.

\paragraph{Policy and training.}
We train a compact DOM-grounded policy with fewer than 10M parameters using PPO. 
At each step, the policy receives the task instruction and the rendered DOM observation, scores the current candidate action set, and executes one DOM-grounded action. 
The policy never observes backend states, completion predicates, or verifier outputs. 
Unless otherwise stated, all policy results are averaged over three random seeds and reported with 95\% confidence intervals.

\paragraph{Environment-construction baselines.}
We compare our verified construction pipeline with five baselines. 
\textsc{No Verification} directly uses raw LLM-generated scaffolds. 
\textsc{Rule-Based} applies deterministic reachability, link, schema, and binding checks. 
\textsc{Single-LLM} uses one GPT-4 verifier prompted to detect all defect categories. 
\textsc{Self-Consistency} samples five independent verifier outputs and applies majority voting. 
\textsc{AutoGen} adapts a general-purpose multi-agent verification framework to the same environment scaffold. 
All methods start from the same scaffolds.

\paragraph{Policy-training baselines.}
To isolate the effect of environment quality and reward design, we train the same compact PPO policy under different synthetic training conditions: raw environments with terminal rewards, raw environments with dense rewards, rule-checked environments with dense rewards, verified environments with terminal rewards, and our full verified environment with state-grounded dense rewards. 
For transfer evaluation, we also compare against GPT-4 direct prompting and GPT-4 with ReAct-style prompting under the same task interface.

\textbf{Metrics.}
For environment quality, we report average defects per environment (\textbf{Def.}), task-blocking defects per environment (\textbf{Block Def.}), the percentage of tasks with at least one bounded executable trace (\textbf{Feasible}), human task success rate (\textbf{Human SR}), state-invariant violation rate after marker-triggered updates (\textbf{State Viol.}), and average curation time per environment (\textbf{Time}). 
For policy learning, we report success rate (\textbf{SR}), average successful-episode length (\textbf{Step}), rejected marker-write rate (\textbf{Reject}), and sample efficiency. 
For simulation efficiency, we report LLM calls, token cost, latency, state fidelity, and training throughput.

\begin{figure}[t]
	\centering
	\includegraphics[width=\linewidth]{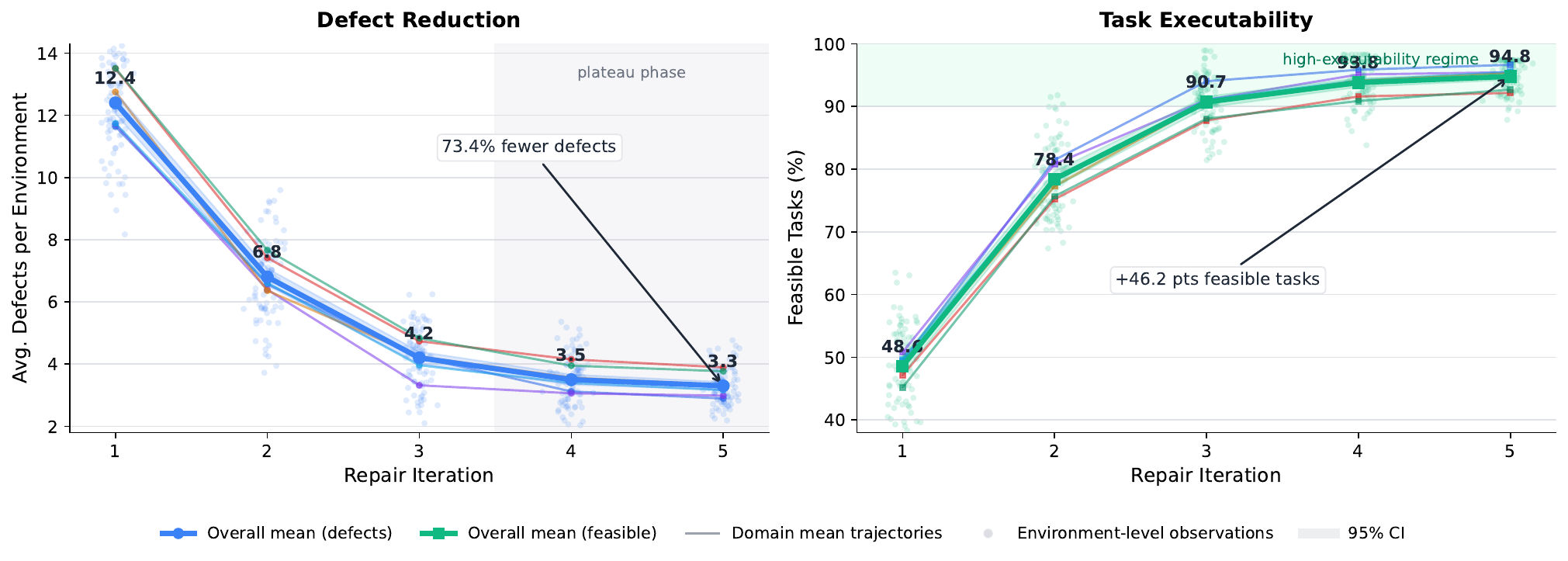}
	\caption{
\textbf{Verification--repair convergence.}
The repair loop rapidly removes defects and increases the fraction of tasks with bounded executable traces. 
The gains saturate after three to four iterations, supporting the use of targeted verification and repair rather than repeated full regeneration.
}
\label{fig:rq1_repair_convergence}
\end{figure}

\subsection{Main results}

\textbf{RQ1: Does verification improve environment executability?}

Table~\ref{tab:rq1_env_quality} shows that raw LLM-generated environments are often invalid, with only 48.6\% of tasks admitting bounded executable traces. Rule-based checks remove simple structural errors but miss task-blocking workflow defects and can reduce human success, while single-pass LLM verification still leaves many blocking defects. In contrast, our method achieves the fewest total and blocking defects, the highest feasible-task rate, the lowest state-violation rate, and lower curation time than self-consistency and AutoGen. 

Fig.~\ref{fig:rq1_repair_convergence} further shows that repair converges quickly: within three iterations, defects drop from 12.4 to 4.2 and feasible tasks rise from 48.6\% to 90.7\%, after which gains saturate.

\begin{table}[t]
\centering
\small
\setlength{\tabcolsep}{4.5pt}
\begin{tabular*}{\linewidth}{@{\extracolsep{\fill}}lcccccc@{}}
\toprule
\textbf{Method} 
& \textbf{Def.$\downarrow$} 
& \textbf{Block Def.$\downarrow$} 
& \textbf{Feasible$\uparrow$} 
& \textbf{Human SR$\uparrow$} 
& \textbf{State Viol.$\downarrow$} 
& \textbf{Time$\downarrow$} \\
\midrule
No Verification     & 12.4 & 4.9 & 48.6\% & 36.9\% & 18.7\% & --    \\
Rule-Based          & 9.8  & 4.1 & 53.2\% & 27.2\% & 14.6\% & 0.5m  \\
Single-LLM          & 6.2  & 2.7 & 70.4\% & 54.3\% & 9.8\%  & 12m   \\
Self-Consistency    & 5.8  & 2.1 & 79.6\% & 69.1\% & 7.2\%  & 35m   \\
AutoGen             & 5.1  & 1.6 & 88.7\% & 89.2\% & 5.5\%  & 28m   \\
\midrule
 \textbf{Ours}       & \textbf{3.3} & \textbf{0.7} & \textbf{94.8\%} & \textbf{91.4\%} & \textbf{2.8\%} & \textbf{18m} \\
\bottomrule
\end{tabular*}
\vspace{3mm}
\caption{
\textbf{Environment executability and fidelity.}
Verification should not merely reduce superficial defects; it should make generated tasks executable under consistent backend dynamics. 
Our method reduces both total defects and task-blocking defects.
}
\label{tab:rq1_env_quality}
\end{table}

\vspace{5mm}

\textbf{RQ2: Does verified synthetic training improve compact policies?}

\begin{figure}[htb]
	\centering
	\includegraphics[width=0.85\linewidth]{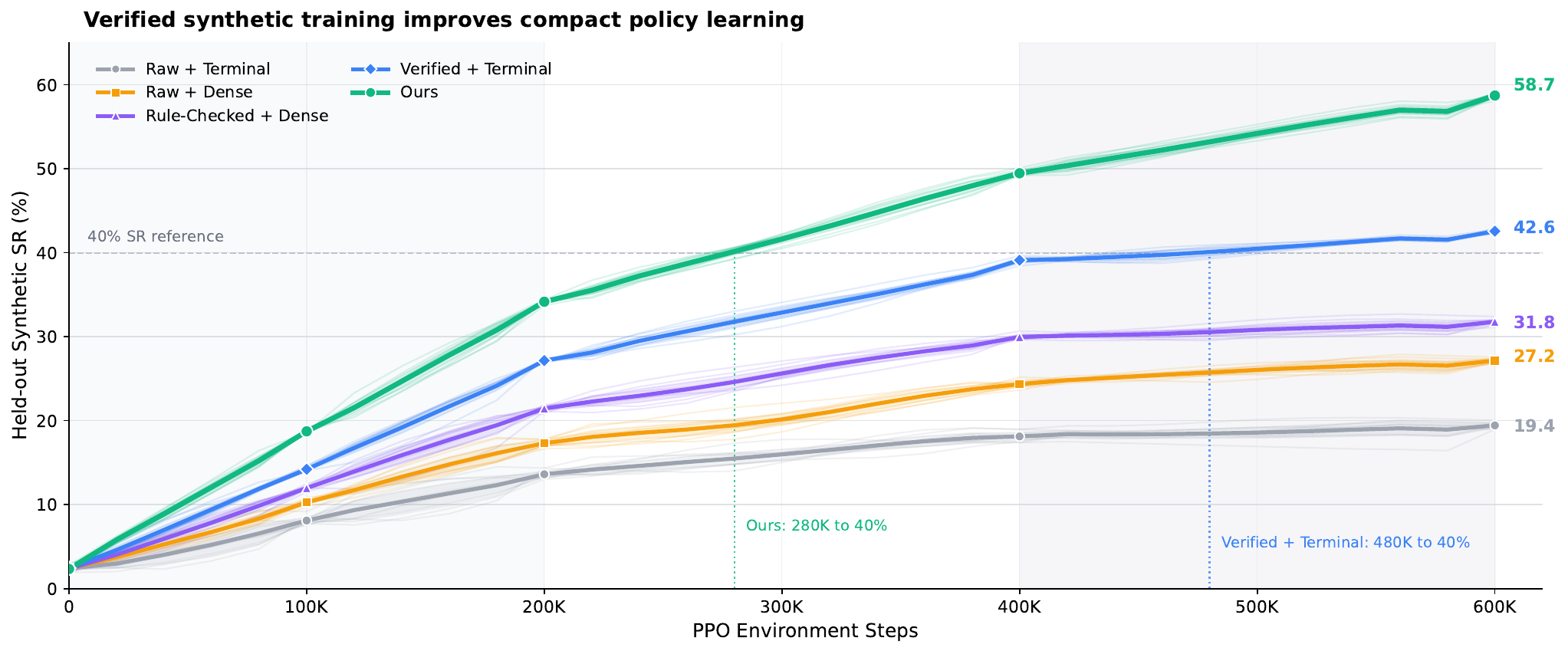}
	\caption{
\textbf{Verified synthetic training improves compact policy learning.}
Raw environments provide noisy supervision, while verified environments make failures attributable to the policy. 
State-grounded dense rewards further accelerate learning and improve final success.
}
\label{fig:rq2_learning_curve}
\end{figure}

We next evaluate whether verified environments improve downstream policy learning. 
Figure~\ref{fig:rq2_learning_curve} compares PPO learning curves on held-out synthetic environments under different training substrates and reward signals. 
Training on raw environments gives weak performance because many failed episodes are caused by environment invalidity rather than policy errors. 
Dense rewards help on raw environments, but the gains remain limited because progress predicates can be noisy when the underlying scaffold is inconsistent. 
Verification alone improves learning under terminal rewards, while combining verified environments with state-grounded dense rewards produces the strongest learning curve and the best final success rate.

\textbf{RQ3: Which components matter?}
\label{sec:rq3_components}

\begin{figure}[H]
    \centering
    \includegraphics[width=0.78\linewidth]{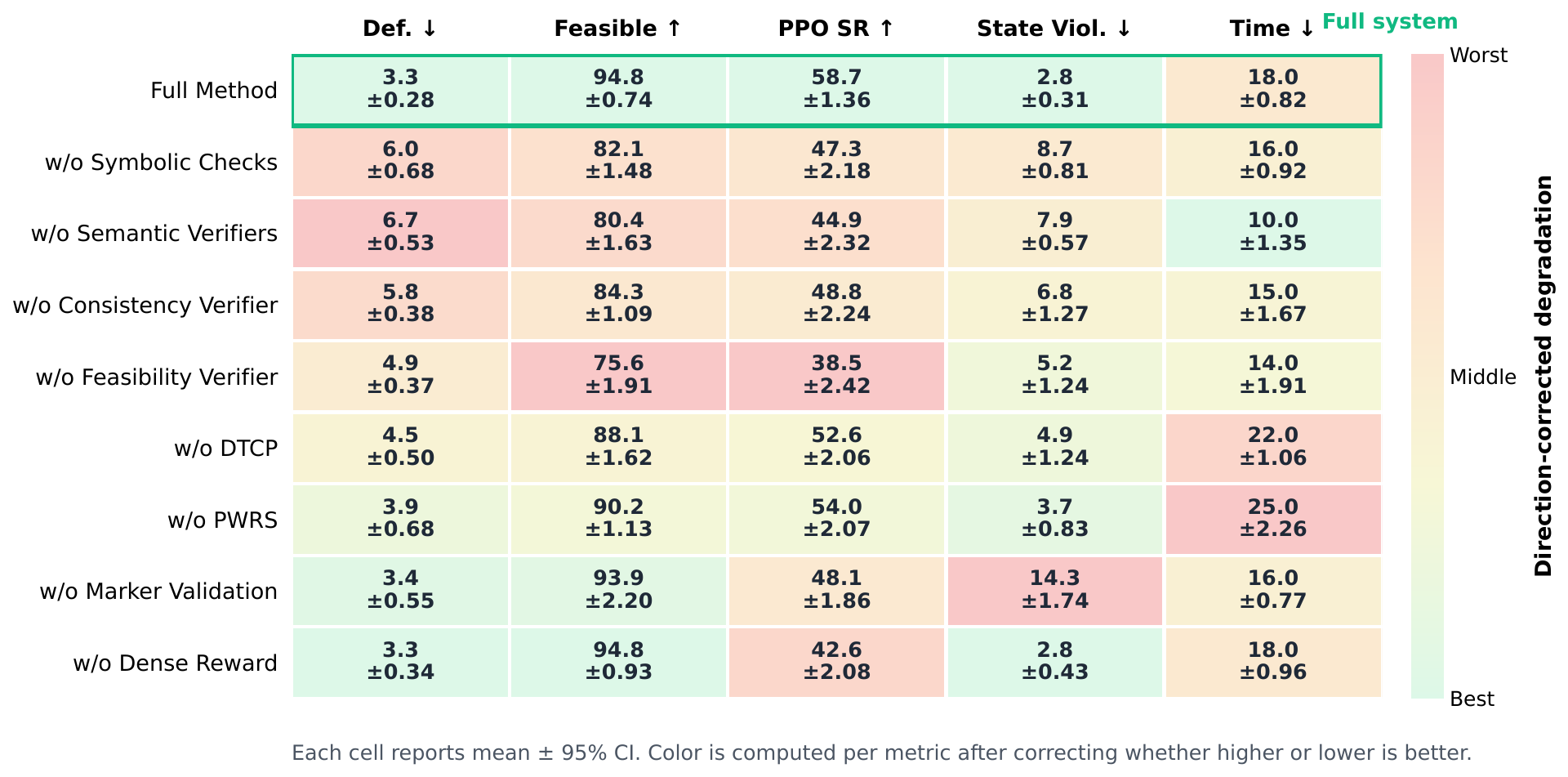}
    \caption{
    \textbf{Component ablation matrix.}
    Rows remove individual components and columns report metrics with 95\% confidence intervals.
    Colors denote direction-corrected degradation.
    Feasibility verification, marker validation, and dense rewards drive executability, state fidelity, and policy learning.
    }
    \label{fig:rq3_component_ablation_matrix}
\end{figure}

We ablate each major component using the same held-out environments and PPO setup. 
Fig.~\ref{fig:rq3_component_ablation_matrix} reports environment quality, task executability, policy success, state violations, and curation cost, with colors indicating direction-corrected degradation. 
The feasibility verifier is most important for executable supervision, as removing it substantially reduces both feasible-task rate and PPO success. 
Marker validation mainly preserves backend fidelity by preventing state violations, while dense rewards primarily improve policy learning without changing environment quality. 
Together, these results show that reliable synthetic web training depends on three complementary ingredients: task-level feasibility verification, state-safe marker updates, and dense state-grounded rewards.

\section{Conclusion}

We studied reinforcement learning for web agents from the perspective of environment reliability, arguing that synthetic websites must be executable, state-consistent, and auditable to provide useful supervision. 
We introduced a verified environment-construction framework that combines offline scaffold verification with event-driven state updates, enabling compact policies to learn from state-grounded synthetic interactions. 
Experiments show that task feasibility verification, marker-level validation, and dense backend-grounded rewards jointly improve environment executability and PPO performance. 
These findings highlight verified environments as a scalable training substrate, with future work extending this paradigm to richer multi-site and multimodal web workflows.

\bibliographystyle{plainnat}
\bibliography{reference}

%%%%%%%%%%%%%%%%%%%%%%%%%%%%%%%%%%%%%%%%%%%%%%%%%%%%%%%%%%%%

\clearpage

\appendix

\section{Additional Technical Details}
\label{app:method_details}

\subsection{Details of the Preliminary Diagnostics}
\label{app:prelim_diagnostic_details}

This subsection provides additional details for the motivating diagnostics in
Section~\ref{sec:preliminaries_diagnostics}. The goal of these diagnostics is
not to serve as the main evaluation, but to empirically justify two design
choices: (i) one-shot generated scaffolds require offline verification, and
(ii) runtime state evolution can be handled by sparse event-driven updates
rather than by calling a generative model at every step.

\subsubsection{Raw Environment Pool}
\label{app:raw_environment_pool}

We construct a diagnostic pool of raw LLM-generated web environments before
any verification or repair. Each environment is generated from a domain-level
site specification and contains page templates, navigation links, database
records, state-change candidates, and task specifications. The environments
cover six representative domains:
\[
\mathcal{D}_{\mathrm{dom}}
=
\{\text{e-commerce},\text{social media},\text{banking},
\text{education},\text{healthcare},\text{government}\}.
\]
For each environment \(E\), we record its page set \(P\), navigation graph
\(L\), database schema \(D\), initialized records, interactive elements, and
task set \(T(E)\). The diagnostic study reports statistics over 500 raw
environments and 6,842 generated tasks.

\subsubsection{Defect Categories and Detection Criteria}
\label{app:defect_detection_criteria}

We use four defect categories. A detected defect is represented as
\[
d=
\langle
\mathrm{type},
\mathrm{loc},
\mathrm{obj},
\mathrm{sev},
\mathrm{ev}
\rangle,
\]
where \(\mathrm{type}\in\{\Delta_S,\Delta_C,\Delta_X,\Delta_T\}\) is the defect
category, \(\mathrm{loc}\) is the affected page, DOM node, database field, or
task, \(\mathrm{obj}\) is the affected object, \(\mathrm{sev}\in[0,1]\) is the
severity score, and \(\mathrm{ev}\) is the supporting evidence.

\begin{table}[h]
\centering
\small
\setlength{\tabcolsep}{4.5pt}
\begin{tabular}{p{2.2cm}p{4.7cm}p{5.3cm}}
\toprule
\textbf{Category} & \textbf{Typical failures} & \textbf{Detection criterion} \\
\midrule
Structural \((\Delta_S)\)
& Broken links, orphaned pages, missing navigation elements
& The page graph is not connected from the entry page, a declared link has no target, or a required UI element is absent. \\

Semantic \((\Delta_C)\)
& Invalid prices, malformed dates, placeholder text, type mismatch
& A rendered value violates the declared schema, expected field type, or page-level content semantics. \\

Consistency \((\Delta_X)\)
& Conflicting prices, inconsistent user attributes, mismatched entity states
& The same entity attribute takes incompatible values across pages, views, or database bindings. \\

Feasibility \((\Delta_T)\)
& Missing buttons, inaccessible forms, incomplete workflow, unsatisfiable task
& No bounded executable trace can satisfy the task completion constraint under the current scaffold. \\
\bottomrule
\end{tabular}
\caption{Defect categories and detection criteria used in the preliminary diagnostics.}
\label{tab:app_defect_criteria}
\end{table}

For each environment \(E\), we count the number of detected defects in category
\(c\) as
\[
N_c(E)=
\sum_{d\in\Delta(E)}
\mathbb{1}[\mathrm{type}(d)=c],
\]
where \(\Delta(E)\) is the full defect set of \(E\).

\subsubsection{Bounded Trace Analysis}
\label{app:bounded_trace_analysis}

To determine whether a task is executable, we perform bounded trace analysis.
Given a task \(t\), an initial state distribution \(\rho_t\), and a completion
constraint \(C_t\), the analysis searches for an action sequence
\[
a_{1:H}=(a_1,\ldots,a_H)
\]
such that
\[
s_0\sim\rho_t,\qquad
s_{k+1}=\mathcal{T}(s_k,a_k),\qquad
a_k\in A(o_k),\qquad
C_t(s_H)=1,
\]
for some horizon \(H\le H_{\max}\). If such a trace exists, the task is
considered executable under the current scaffold. Otherwise, the task is marked
as blocked. This trace search is used only for diagnostics, not for policy
training.

\begin{algorithm}[t]
\caption{Bounded Trace Analysis for Task Feasibility}
\label{alg:bounded_trace_analysis}
\small
\begin{algorithmic}[1]
\REQUIRE Environment \(E\), task \(t\), initial states \(\mathcal S_0\), horizon \(H_{\max}\)
\ENSURE Executable flag and blocking evidence
\STATE Initialize queue \(Q\leftarrow\{(s_0,[]):s_0\in\mathcal S_0\}\)
\STATE Initialize visited set \(\mathcal V\leftarrow\emptyset\)
\FOR{\(h=0,\ldots,H_{\max}\)}
    \STATE Initialize next queue \(Q'\leftarrow\emptyset\)
    \FOR{each \((s,\pi)\in Q\)}
        \IF{\(C_t(s)=1\)}
            \RETURN \((\textsc{Executable},\pi,\emptyset)\)
        \ENDIF
        \STATE \(o\leftarrow\mathrm{Render}(s)\)
        \STATE Extract candidate actions \(A(o)\)
        \FOR{each \(a\in A(o)\)}
            \STATE Execute \(s'\leftarrow\mathcal T(s,a)\)
            \STATE Record transition evidence \(\mathrm{ev}(s,a,s')\)
            \IF{\(s'\notin\mathcal V\)}
                \STATE \(Q'\leftarrow Q'\cup\{(s',\pi\circ a)\}\)
                \STATE \(\mathcal V\leftarrow\mathcal V\cup\{s'\}\)
            \ENDIF
        \ENDFOR
    \ENDFOR
    \STATE \(Q\leftarrow Q'\)
\ENDFOR
\STATE Extract blocking evidence from failed expansions
\RETURN \((\textsc{Blocked},\emptyset,\mathrm{ev})\)
\end{algorithmic}
\end{algorithm}

\subsubsection{Blocking Attribution}
\label{app:blocking_attribution}

When a task is blocked, we attribute the failure to the earliest defect category
that prevents progress along the attempted workflow. Let
\(\mathrm{Reach}_h(E,t)\) denote the set of states reachable within \(h\) steps
from the task initial state. A defect \(d\) is considered task-blocking if it
invalidates all outgoing progress transitions from the current reachable
frontier. We define
\[
\operatorname{Block}_c(E,t)=1
\]
if at least one defect of category \(c\) is responsible for blocking all bounded
traces for task \(t\).

If multiple defect categories are implicated, we apply the following priority
order based on causal proximity to execution:
\[
\Delta_T \succ \Delta_S \succ \Delta_X \succ \Delta_C .
\]
Feasibility defects are assigned first because they directly indicate missing
workflow steps or inaccessible required controls. Structural defects are next
because they prevent reaching necessary pages. Consistency and semantic defects
are assigned when the task path exists but the required state or content is
invalid.

The task-blocking rate reported in the main text is
\[
B_c(E)=
\frac{1}{|T(E)|}
\sum_{t\in T(E)}
\mathbb{1}\!\left[\operatorname{Block}_c(E,t)=1\right].
\]
The global blocking rate for category \(c\) across environments is
\[
\overline{B}_c
=
\frac{1}{|\mathcal E|}
\sum_{E\in\mathcal E}
B_c(E).
\]

\subsubsection{State-Change Markers and Runtime Statistics}
\label{app:marker_runtime_statistics}

A state-change marker identifies a UI event that may modify persistent backend
state. Each marker is represented as
\[
m=(e_m,\mathrm{pre}_m,R_m,W_m,\Omega_m),
\]
where \(e_m\) is the triggering element, \(\mathrm{pre}_m\) is a precondition,
\(R_m\) and \(W_m\) are the permitted read and write fields, and \(\Omega_m\)
is the operation signature. Typical marker-triggered events include
\texttt{add-to-cart}, \texttt{submit-order}, \texttt{update-profile},
\texttt{create-post}, and \texttt{submit-application}.

For a successful trajectory
\[
\tau=(s_0,a_0,s_1,a_1,\ldots,s_H),
\]
we compute the marker-trigger rate
\[
M(\tau)=
\frac{1}{H}
\sum_{k=0}^{H-1}
\mathbb{1}\!\left[\mu(p_k,a_k)\neq\varnothing\right],
\]
and the number of runtime state-write calls
\[
U(\tau)=
\sum_{k=0}^{H-1}
\mathbb{1}\!\left[\mu(p_k,a_k)\neq\varnothing\right].
\]
Here, \(\mu(p_k,a_k)\) returns the marker triggered by action \(a_k\) on page
\(p_k\), if such a marker exists. Deterministic UI actions, including
navigation, scrolling, menu expansion, and local text entry, do not trigger
persistent state writes.

For marker-triggered operations, a candidate state delta \(\hat{\delta}_k\) is
accepted only if it satisfies the marker precondition, schema constraints, and
environment invariants:
\[
\delta_k=
\mathrm{Validate}_{m_k}(\hat{\delta}_k;s_k,a_k)
\in
\mathcal U_{m_k}(s_k,a_k)\cup\{\bot\}.
\]
The accepted-delta rate is
\[
A(\tau)=
\frac{
\sum_{k=0}^{H-1}
\mathbb{1}\!\left[\mu(p_k,a_k)\neq\varnothing\right]
\mathbb{1}\!\left[\delta_k\neq\bot\right]
}{
\sum_{k=0}^{H-1}
\mathbb{1}\!\left[\mu(p_k,a_k)\neq\varnothing\right]+\epsilon
},
\]
where \(\epsilon\) is a small constant used only to avoid division by zero for
trajectories with no marker-triggered event.

\subsubsection{Confidence Intervals and Visualization Protocol}
\label{app:diagnostic_visualization}

For each scalar statistic \(z_1,\ldots,z_n\), we report the empirical mean and
the normal-approximation 95\% confidence interval:
\[
\bar z=\frac{1}{n}\sum_{i=1}^n z_i,
\qquad
\mathrm{CI}_{95}
=
1.96\frac{\mathrm{std}(z_1,\ldots,z_n)}{\sqrt n}.
\]
In Fig.~\ref{fig:prelim_diagnostics}, each point corresponds to one environment
or one trajectory, depending on the panel. Jitter is used only for visualization
and does not affect the reported statistics. Panel (a) shows per-environment
defect counts. Panel (b) shows task-blocking rates conditioned on detected
defects. Panel (c) shows marker-trigger rates grouped by domain. Panel (d)
shows the relationship between episode length and the number of runtime
state-write calls, with the fitted line used only as a visual summary.

These diagnostics are intended to motivate the method design. The main
experimental evaluation in Section~\ref{sec:experiments} separately measures
environment quality, repair effectiveness, and downstream policy performance.

\subsection{Canonical Parsing and Symbolic Verification}
\label{app:canonical_verification}

\paragraph{Raw scaffold artifacts.}
Given a site description \(q\), the generator produces raw artifacts
\[
\widetilde{\mathcal E}
=
(\widetilde{\mathcal P},
 \widetilde{\mathcal L},
 \widetilde{\mathcal D},
 \widetilde{\mathcal Q}),
\]
where \(\widetilde{\mathcal P}\) contains page templates, \(\widetilde{\mathcal L}\) contains declared links and menus, \(\widetilde{\mathcal D}\) contains database schema and initialized records, and \(\widetilde{\mathcal Q}\) contains task instructions with completion constraints. The parser converts these artifacts into the canonical scaffold \(\mathcal E^{(0)}\) by extracting DOM nodes, interactive elements, database bindings, and state-change markers.

\paragraph{Page and binding representation.}
Each page \(p\in\mathcal P\) is represented as
\[
p=(\tau_p,\mathcal I_p,\Gamma_p),
\]
where \(\tau_p\) is a DOM template, \(\mathcal I_p\) is the set of interactive elements, and \(\Gamma_p\) maps DOM variables to database fields. For an interactive element \(e\in\mathcal I_p\), the parser records its element type, admissible arguments, page-level effect, and marker association.

\paragraph{Marker schema.}
A marker \(m=(e_m,\mathrm{pre}_m,R_m,W_m,\Omega_m)\) is attached to a persistent state-changing element. The precondition \(\mathrm{pre}_m\) specifies when the operation is valid. The sets \(R_m\) and \(W_m\) specify database fields that the operation reads and writes. The operation signature \(\Omega_m\) contains the operation name, target entity type, argument schema, output delta schema, and invariant checks.

\paragraph{Symbolic checks.}
The symbolic verifier computes
\[
\Delta_{\mathrm{sym}}
=
\Delta_{\mathrm{reach}}
\cup
\Delta_{\mathrm{link}}
\cup
\Delta_{\mathrm{schema}}
\cup
\Delta_{\mathrm{bind}}
\cup
\Delta_{\mathrm{mark}}.
\]
The individual defect sets are
\[
\begin{aligned}
\Delta_{\mathrm{reach}}
&=
\{p\in\mathcal P:
p\notin\mathrm{Reach}(p_{\mathrm{home}},\mathcal L)\},\\
\Delta_{\mathrm{link}}
&=
\{(p_i,p_j)\in\mathcal L:
p_j\notin\mathcal P\},\\
\Delta_{\mathrm{schema}}
&=
\{g:
g\ \text{violates a type, key, range, or required-field constraint in}\ \mathcal D\},\\
\Delta_{\mathrm{bind}}
&=
\{(p,g):
g\in\Gamma_p,\;
\mathrm{field}(g)\notin\mathrm{Fields}(\mathcal D)\},\\
\Delta_{\mathrm{mark}}
&=
\{m\in\mathcal M:
R_m\cup W_m\nsubseteq\mathrm{Fields}(\mathcal D)
\ \text{or}\ 
\Omega_m\ \text{has an invalid argument schema}\}.
\end{aligned}
\]
Here, \(p_{\mathrm{home}}\) is the entry page, \(\mathrm{Reach}\) returns reachable pages under the navigation graph, and \(\mathrm{Fields}(\mathcal D)\) is the set of valid database fields. These deterministic checks are run before semantic verification and after each repair iteration on the affected subgraph.

\subsection{Semantic Verifiers and Defect-Triggered Coordination}
\label{app:verifier_coordination}

\paragraph{Verifier roles.}
We use four semantic verifiers:
\[
\mathcal J=
\{\mathrm{str},\mathrm{sem},\mathrm{cons},\mathrm{feas}\}.
\]
The structural verifier checks layout coherence, missing required elements, and navigation anomalies beyond link existence. The semantic verifier checks content validity, content--type compatibility, placeholder text, and implausible values. The consistency verifier checks entity attributes across pages and rendered states. The feasibility verifier searches for bounded executable traces that satisfy task constraints.

\paragraph{Structured defect report.}
Each verifier \(j\in\mathcal J\) returns a set of reports
\[
\mathcal A_j(\mathcal E^{(r)})
\rightarrow
\Delta_j^{(r)}
=
\{d_{j,1}^{(r)},\ldots,d_{j,n_j}^{(r)}\}.
\]
Each report has the schema
\[
d=
\langle
\mathrm{type},
\mathrm{loc},
\mathrm{obj},
\mathrm{sev},
\mathrm{ev},
c
\rangle,
\]
where \(\mathrm{type}\) is the defect category, \(\mathrm{loc}\) is the page, DOM node, database field, marker, or task where the defect occurs, \(\mathrm{obj}\) is the affected object, \(\mathrm{sev}\in[0,1]\) is severity, \(\mathrm{ev}\) is supporting evidence, and \(c\in[0,1]\) is verifier confidence.

\paragraph{Deduplication.}
Reports are merged by a canonical key
\[
\kappa(d)=
(\mathrm{type}(d),\mathrm{loc}(d),\mathrm{obj}(d)).
\]
All reports with the same key are grouped into one defect candidate. The main text defines the aggregated confidence in Eq.~\eqref{eq:defect_conf}.

\paragraph{Defect-triggered coordination.}
Verifier communication is routed by defect type:
\[
B(\mathrm{type}(d))\subseteq\mathcal J,
\]
where \(B\) returns verifiers that need to re-check downstream effects of \(d\). We use the following routing rules:
\[
\begin{aligned}
B(\mathrm{structural}) &= \{\mathrm{feas}\},\\
B(\mathrm{semantic}) &= \{\mathrm{cons}\},\\
B(\mathrm{consistency}) &= \{\mathrm{feas}\},\\
B(\mathrm{feasibility}) &= \{\mathrm{str},\mathrm{cons}\}.
\end{aligned}
\]
For a defect \(d\), the coordinator adds targeted requests
\[
Q \leftarrow Q\cup\{(d,j'):j'\in B(\mathrm{type}(d))\}.
\]
The receiving verifier checks only the affected pages, database fields, markers, and tasks referenced in the report evidence.

\paragraph{Accepted defect set.}
A defect is accepted for repair when
\[
\mathrm{conf}(d)\ge\tau_c
\quad
\text{or}
\quad
\mathrm{sev}(d)\ge\tau_s,
\]
where \(\tau_c\) is the confidence threshold and \(\tau_s\) is the severity threshold. This rule preserves high-severity single-verifier defects while filtering low-confidence reports.

\subsection{Repair Operators and Termination}
\label{app:repair_ops}

\paragraph{Dependency graph.}
Accepted defects form a dependency graph
\[
G_\Delta^{(r)}=
(\Delta_{\mathrm{acc}}^{(r)},\mathcal R_\Delta^{(r)}),
\]
where \((d_i,d_j)\in\mathcal R_\Delta^{(r)}\) means that \(d_i\) must be repaired before \(d_j\). Dependencies are added when a database schema defect affects page rendering, when a marker defect affects task feasibility, and when a structural defect blocks access to pages used by semantic and consistency checks.

\paragraph{Ready set.}
At iteration \(r\), the ready set is
\[
\mathrm{Ready}^{(r)}
=
\{d\in\Delta_{\mathrm{acc}}^{(r)}:
\mathrm{Pred}_{G_\Delta}(d)\subseteq\mathcal S_{\mathrm{done}}^{(r)}\},
\]
where \(\mathrm{Pred}_{G_\Delta}(d)\) is the predecessor set of defect \(d\), and \(\mathcal S_{\mathrm{done}}^{(r)}\) is the set of repaired defects. The selected defect is
\[
d^\star=
\arg\max_{d\in\mathrm{Ready}^{(r)}}\mathrm{score}(d),
\]
where \(\mathrm{score}(d)\) is defined in Eq.~\eqref{eq:repair_score}.

\paragraph{Repair operators.}
The repair operator \(\mathcal R_{d^\star}\) edits only objects referenced by the accepted report:
\[
\mathcal E^{(r+1)}
=
\mathcal R_{d^\star}(\mathcal E^{(r)}).
\]
We use the following repair families.

\begin{itemize}[leftmargin=*]
    \item \textbf{Structural repair.} Adds missing pages, repairs invalid links, restores required navigation elements, and reconnects orphaned pages to the navigation graph.
    \item \textbf{Semantic repair.} Rewrites invalid content while preserving database bindings and task-relevant entities.
    \item \textbf{Consistency repair.} Selects a canonical value from the database and propagates it to all dependent DOM bindings.
    \item \textbf{Feasibility repair.} Adds missing form fields, buttons, intermediate pages, marker signatures, and database writes required by a bounded task trace.
    \item \textbf{Marker repair.} Corrects marker preconditions, argument schema, read--write sets, and invariant checks.
\end{itemize}

\paragraph{Incremental re-verification.}
After repairing \(d^\star\), we re-check the affected subgraph
\[
\mathcal N(d^\star)
=
\{p:\mathrm{dist}_{\mathcal L}(p,\mathrm{loc}(d^\star))\le 1\}
\cup
\mathrm{Tasks}(d^\star)
\cup
\mathrm{Markers}(d^\star),
\]
where \(\mathrm{dist}_{\mathcal L}\) is graph distance in the navigation graph. This prevents each repair iteration from re-running all verifiers on the entire environment.

\paragraph{Feasibility termination.}
For a task \(t\), feasibility is accepted when the feasibility verifier finds a bounded executable trace
\[
\pi_{1:H}=(a_1,\ldots,a_H)
\quad
\text{such that}
\quad
s_{k+1}=F(s_k,a_k),\; a_k\in\mathcal A(o_k),\; C_t(s_H)=1.
\]
The repair loop terminates when all accepted critical defects are resolved and every task has at least one such bounded trace.

\subsection{Event-Driven Simulator Details}
\label{app:simulator_details}

\paragraph{Deterministic transition.}
The deterministic transition \(F_{\mathrm{det}}\) handles navigation, local input editing, scroll state, menu expansion, client-side field validation, and page rendering. It does not modify persistent database records outside session-local variables.

\paragraph{Candidate delta generation.}
For a marker-triggered action, the state writer receives
\[
z_k=
\mathrm{Pack}(p_k,b_{k,R_m},\sigma_k,a_k,\Omega_m),
\]
where \(b_{k,R_m}\) is the subset of database records referenced by the marker read set. The state writer outputs a candidate delta
\[
\hat{\delta}_k=G_\psi(z_k).
\]
The output schema contains only field-level writes, session updates, and a short justification tied to \(\Omega_m\). Full-page natural-language regeneration is not accepted as a state update.

\paragraph{Feasible update set.}
For marker \(m\), the feasible update set is
\[
\mathcal U_m(s_k,a_k)
=
\left\{
\delta:
\mathrm{pre}_m(s_k,a_k)=1,\;
\mathrm{RW}(\delta)\subseteq R_m\cup W_m,\;
b_k\oplus\delta\models\mathcal D,\;
\mathrm{Inv}(s_k\oplus\delta)=1
\right\}.
\]
Here, \(\mathrm{RW}(\delta)\) is the set of fields read and written by the delta, \(b_k\oplus\delta\) is the database after applying the delta, and \(\mathrm{Inv}\) contains environment invariants including schema validity, entity identity consistency, permission constraints, and task-independent workflow constraints.

\paragraph{Validation.}
The validation operator returns
\[
\mathrm{Validate}_m(\hat{\delta}_k;s_k,a_k)
=
\begin{cases}
\hat{\delta}_k,
& \hat{\delta}_k\in\mathcal U_m(s_k,a_k),\\
\bot,
& \hat{\delta}_k\notin\mathcal U_m(s_k,a_k).
\end{cases}
\]
If the result is \(\bot\), the simulator returns the original state, sets \(\epsilon_k=1\), and exposes the rejected operation in the transition log. Otherwise, the simulator applies \(s_{k+1}=s_k\oplus\delta_k\), re-renders the current page through \(\mathrm{Render}\), and records the state delta for reward computation and debugging.

\subsection{Reward Compilation and PPO Training}
\label{app:ppo_details}

\paragraph{Predicate compilation.}
Each completion constraint \(C_t\) is decomposed into predicates
\[
\Phi_t=\{\phi_{t,1},\ldots,\phi_{t,M_t}\}.
\]
Predicates are implemented as database queries, session-state checks, and DOM selectors. Examples include checking that a required page has been visited, a field has a valid value, a target entity has been created, an attribute has been updated correctly, and a rendered page reflects the backend value.

\paragraph{Candidate action scoring.}
The policy receives the task instruction \(x_t\), the rendered observation \(o_k\), and the candidate action set \(\mathcal A(o_k)\). It scores each candidate action by
\[
\ell_\theta(a;o_k,x_t),
\quad a\in\mathcal A(o_k),
\]
and normalizes over the current candidate set:
\[
\pi_\theta(a\mid o_k,x_t)
=
\frac{
\exp(\ell_\theta(a;o_k,x_t))
}{
\sum_{a'\in\mathcal A(o_k)}
\exp(\ell_\theta(a';o_k,x_t))
}.
\]
Since sampling is restricted to \(\mathcal A(o_k)\), invalid DOM actions do not appear in the policy distribution. Rejected transitions in the main reward arise from marker validation failures, not from selecting unavailable DOM elements.

\paragraph{Trajectory collection.}
For each task \(t\), PPO collects rollouts
\[
\tau=
(o_0,a_0,r_0,o_1,\ldots,o_H)
\]
by executing \(\pi_\theta\) in the event-driven simulator. The environment computes rewards using Eq.~\eqref{eq:reward}; the policy does not observe the backend state, completion constraint, or progress predicates.

\paragraph{Advantage estimation.}
We use generalized advantage estimation:
\[
\hat A_k
=
\sum_{\ell=0}^{H-k-1}
(\xi\lambda_{\mathrm{GAE}})^\ell
\delta^V_{k+\ell},
\qquad
\delta^V_k
=
r_k+\xi V_{\theta_{\mathrm{old}}}(o_{k+1},x_t)
-
V_{\theta_{\mathrm{old}}}(o_k,x_t),
\]
where \(\xi\) is the RL discount factor, \(\lambda_{\mathrm{GAE}}\) controls the bias--variance trade-off, and \(V_\theta\) is the value function.

\paragraph{PPO objective.}
The probability ratio is
\[
\rho_k(\theta)
=
\frac{
\pi_\theta(a_k\mid o_k,x_t)
}{
\pi_{\theta_{\mathrm{old}}}(a_k\mid o_k,x_t)
}.
\]
The policy is optimized with the clipped objective
\[
\begin{aligned}
\mathcal L_{\mathrm{PPO}}(\theta)
=&
\mathbb E_k
\left[
\min\left(
\rho_k(\theta)\hat A_k,
\mathrm{clip}(\rho_k(\theta),1-\epsilon,1+\epsilon)\hat A_k
\right)
\right] \\
&-
c_v\mathbb E_k
\left[
\left(V_\theta(o_k,x_t)-\hat R_k\right)^2
\right]
+
c_h\mathbb E_k
\left[
\mathcal H(\pi_\theta(\cdot\mid o_k,x_t))
\right].
\end{aligned}
\]
where \(\epsilon\) is the clipping threshold, \(\hat R_k\) is the empirical return, \(c_v\) weights the value loss, \(c_h\) weights the entropy bonus, and \(\mathcal H\) is policy entropy.

\paragraph{Evaluation-time policy.}
At evaluation time, the learned policy uses only \(o_k\), \(x_t\), and the DOM-grounded candidate actions \(\mathcal A(o_k)\). It does not call the generation model, semantic verifiers, repair operators, or constrained state writer.

\subsection{Four agents design}
\label{agent-sepcific}
Four specialized LLM-based agents collaborate to detect complex defects requiring semantic understanding. Each agent is implemented as a GPT-4 instance with domain-specific system prompts and structured output schemas.

\textbf{Structure Validator (SV):} Analyzes page layouts and navigation flows to identify structural anomalies beyond simple link checking. SV examines: (1) navigation consistency (breadcrumb accuracy, menu completeness), (2) page hierarchy adherence (category $\rightarrow$ subcategory $\rightarrow$ item), and (3) required element presence (search bars, footers, headers). Detection is based on embedding similarity using a fine-tuned sentence transformer to identify outlier page structures.

\textbf{Content Auditor (CA):} Examines content for semantic validity using GPT-4's world knowledge. CA detects: (1) placeholder text patterns (``Lorem ipsum'', ``[TODO]'', ``Example''), (2) content-metadata mismatches (product description contradicting title), (3) implausible content (negative prices, future birthdates), and (4) inappropriate content for page type (technical jargon on consumer pages).

\textbf{Consistency Checker (CC):} Cross-references data across pages using an entity database. For each entity $e$ with attributes $\{a_1, \ldots, a_k\}$, CC tracks all occurrences across pages and flags when $a_i(p_j) \neq a_i(p_k)$ for any attribute. Semantic equivalence is handled via embedding similarity (threshold 0.92) for string attributes.

\textbf{Task Feasibility Analyzer (TFA):} Traces task execution paths by simulating agent trajectories. For each task $t \in \mathcal{T}$, TFA generates an action sequence using GPT-4 and verifies: (1) all required pages exist, (2) all form fields are accessible, (3) all buttons are clickable, and (4) the goal state is reachable.

\begin{table}[htbp]
\begin{center}
\scriptsize
\setlength{\tabcolsep}{2pt}
\begin{tabularx}{0.9\linewidth}{p{1.6cm}X p{2.6cm}}
\toprule
\textbf{Type} & \textbf{Example} & \textbf{Detection} \\
\midrule
Structural & Link to \texttt{/checkout} returns 404 & Graph + HTTP \\
Semantic & Price field shows ``TBD'' instead of \$29.99 & Schema + LLM \\
Consistency & Product \$49 on list, \$59 on detail & Cross-page DB \\
Feasibility & ``Add to cart'' button not clickable & Task simulation \\
\bottomrule
\end{tabularx}
\end{center}
\caption{Defect Examples and Detection Complexity}
\label{tab:defect_examples}
\end{table}

\subsubsection{Defect-Triggered Communication Protocol}
\label{DTCP}
Rather than independent parallel execution, agents communicate through DTCP to enable cross-agent defect correlation. When agent $A_i$ detects a defect $d$, it broadcasts a structured message to relevant agents:

\begin{equation}
\text{msg}(d) = \langle \text{type}, \text{location}, \text{severity}, \text{evidence} \rangle
\end{equation}

Receiving agents use domain-specific rules to determine follow-up actions:
\begin{itemize}
\item $SV \rightarrow TFA$: Structural defects trigger task re-verification
\item $CA \rightarrow CC$: Semantic issues trigger consistency checks
\item $CC \rightarrow TFA$: Data conflicts trigger feasibility re-analysis
\item $TFA \rightarrow SV$: Workflow gaps trigger navigation review
\end{itemize}

% This selective communication reduces API calls compared to full broadcast while improving recall by 8.2\% through targeted follow-up verification.
% This selective communication reduces API calls by 34\% compared to full broadcast while improving recall by 8.2\% through targeted follow-up verification.

\subsection{Agent Configuration Details}
\label{app:agent_config}

\paragraph{Structural Verification Agent}

The Structural Verification Agent ($\mathcal{A}_S$) employs the following prompt template:
\begin{figure}[htbp]
    \centering
    \includegraphics[width=1\linewidth]{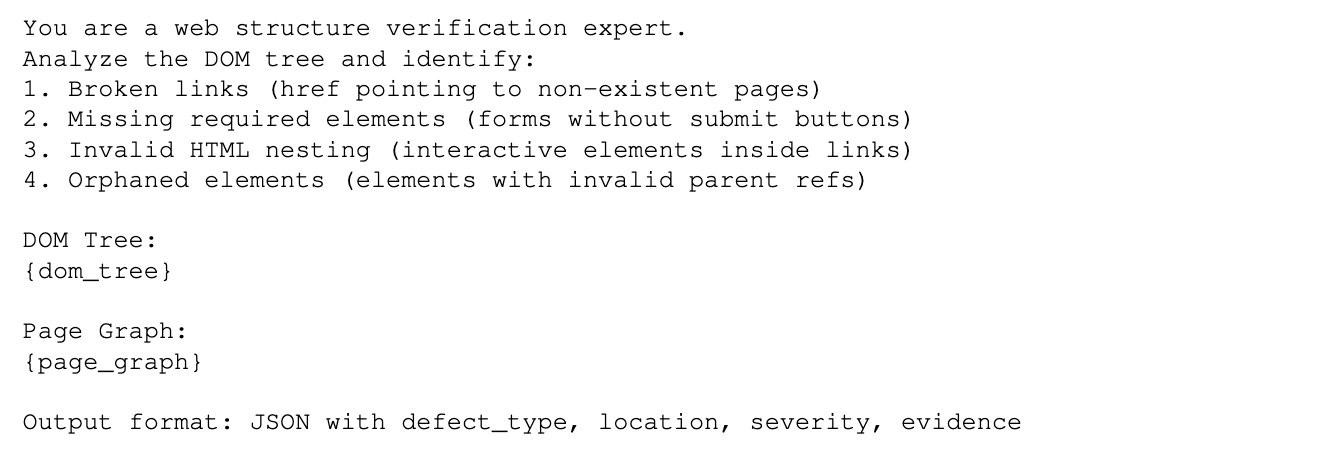}
    \caption{Structural Verification Agent}
    \label{fig:SV}
\end{figure}

\paragraph{Consistency Verification Agent}

The Consistency Verification Agent ($\mathcal{A}_C$) uses:
\begin{figure}[htbp]
    \centering
    \includegraphics[width=1\linewidth]{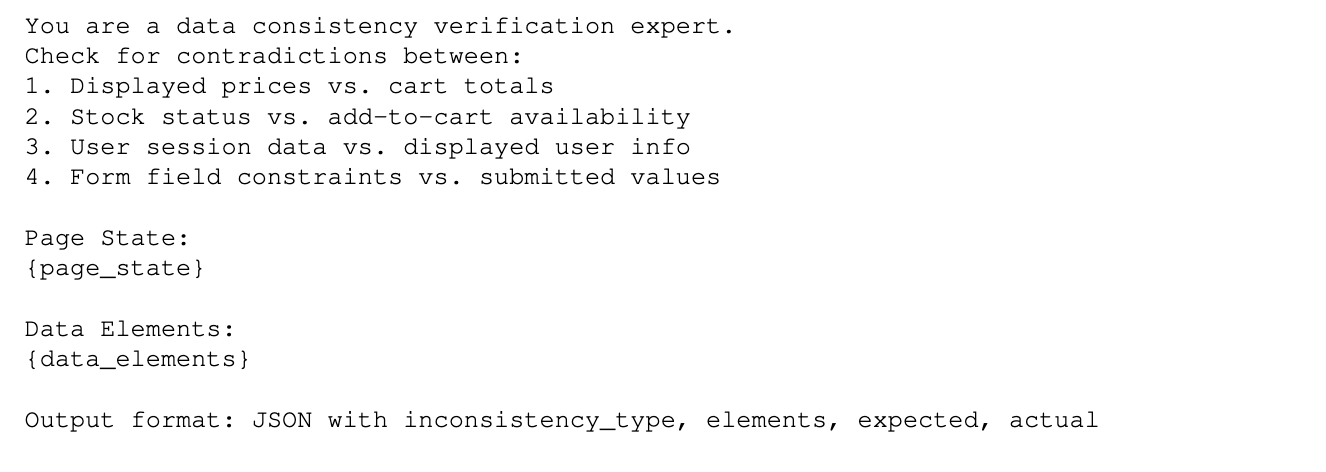}
    \caption{Consistency Verification Agent}
    \label{fig:CV}
\end{figure}

\paragraph{Cross-Page Verification Agent}

The Cross-Page Verification Agent ($\mathcal{A}_X$) configuration:
\begin{figure}[htbp]
    \centering
    \includegraphics[width=1\linewidth]{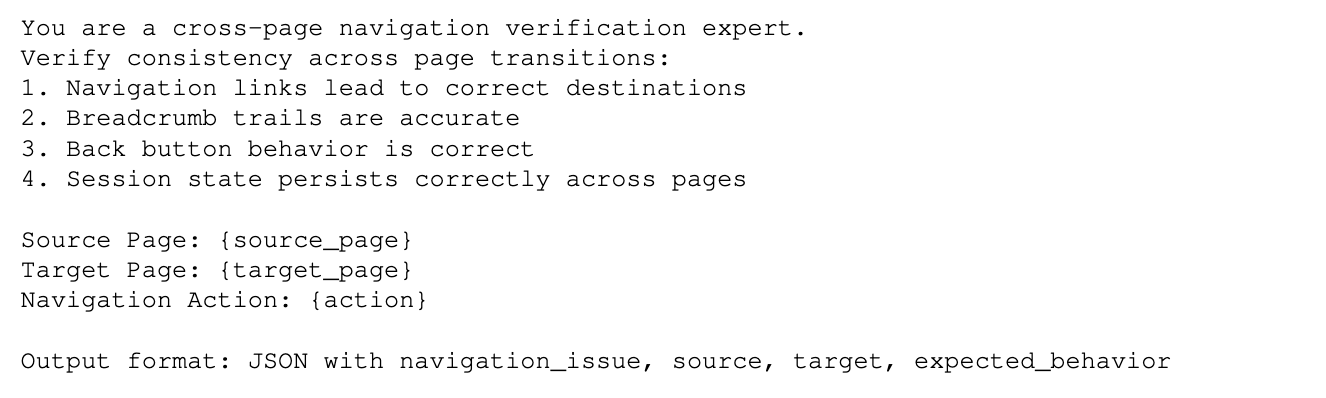}
    \caption{Cross-Page Verification Agent}
    \label{fig:Cross-Page}
\end{figure}

\paragraph{Task Flow Verification Agent}

The Task Flow Verification Agent ($\mathcal{A}_T$) prompt:
\begin{figure}[htbp]
    \centering
    \includegraphics[width=\linewidth]{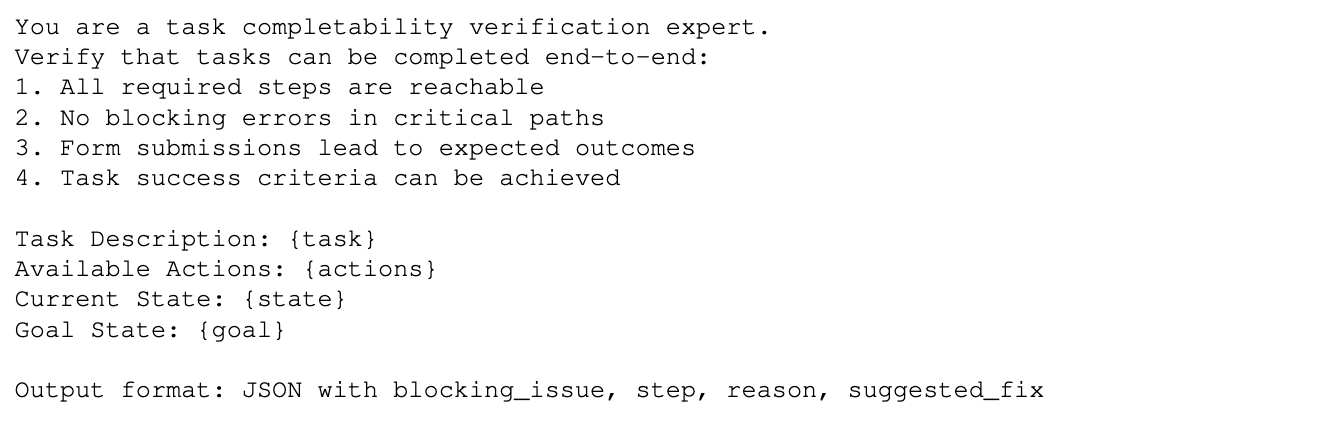}
    \caption{Task Flow Verification Agent}
    \label{fig:task-agent}
\end{figure}

\subsection{DTCP Protocol Specification}
\label{app:dtcp}

Algorithm~\ref{alg:dtcp} presents the complete Defect-Triggered Communication Protocol.

\begin{algorithm}[htbp]
\caption{Defect-Triggered Communication Protocol (DTCP)}
\label{alg:dtcp}
\small
\begin{algorithmic}[1]
\REQUIRE Agents $\{\mathcal{A}_S, \mathcal{A}_C, \mathcal{A}_X, \mathcal{A}_T\}$, Environment $\mathcal{E}$, Thresholds $\{\tau_1, \tau_2\}$
\ENSURE Aggregated defect reports $\Delta$
\STATE Initialize message queue $Q \leftarrow \emptyset$
\STATE Initialize defect set $\Delta \leftarrow \emptyset$
\FOR{each agent $\mathcal{A}_i$ in parallel}
    \STATE $\delta_i \leftarrow \mathcal{A}_i.\textsc{Verify}(\mathcal{E})$
    \FOR{each defect $d \in \delta_i$}
        \IF{$d.\text{confidence} < \tau_1$}
            \STATE $Q.\textsc{Push}(\textsc{RequestVerification}(d, \mathcal{A}_i))$
        \ELSIF{$d.\text{affects\_other\_domain}$}
            \STATE $Q.\textsc{Push}(\textsc{NotifyRelated}(d, \mathcal{A}_i))$
        \ENDIF
        \STATE $\Delta \leftarrow \Delta \cup \{d\}$
    \ENDFOR
\ENDFOR
\WHILE{$Q \neq \emptyset$}
    \STATE $msg \leftarrow Q.\textsc{Pop}()$
    \STATE $\mathcal{A}_j \leftarrow \textsc{GetTargetAgent}(msg)$
    \STATE $response \leftarrow \mathcal{A}_j.\textsc{ProcessMessage}(msg)$
    \IF{$response.\text{type} = \textsc{Confirm}$}
        \STATE $\Delta[msg.defect].\text{confidence} \mathrel{+}= \tau_2$
    \ELSIF{$response.\text{type} = \textsc{Reject}$}
        \STATE $\Delta.\textsc{Remove}(msg.defect)$
    \ELSIF{$response.\text{type} = \textsc{NewDefect}$}
        \STATE $\Delta \leftarrow \Delta \cup \{response.defect\}$
    \ENDIF
\ENDWHILE
\RETURN $\Delta$
\end{algorithmic}
\end{algorithm}

\subsection{PWRS Algorithm Details}
\label{app:pwrs}

The Priority-Weighted Repair Scheduling (PWRS) algorithm considers defect dependencies and repair costs.

\begin{algorithm}[ht]
\caption{Priority-Weighted Repair Scheduling (PWRS)}
\label{alg:pwrs}
\small
\begin{algorithmic}[1]
\REQUIRE Defects $\Delta$, Dependency graph $G_d$, Weight function $w$
\ENSURE Ordered repair schedule $S$
\STATE $S \leftarrow []$
\STATE Compute priority scores: $\forall d \in \Delta: p(d) = w(d) \cdot \text{downstream}(d, G_d)$
\STATE Build dependency-aware priority queue $PQ$ ordered by $p(d)$
\WHILE{$PQ \neq \emptyset$}
    \STATE $d \leftarrow PQ.\textsc{PopMax}()$
    \IF{$\textsc{DependenciesSatisfied}(d, S, G_d)$}
        \STATE $S.\textsc{Append}(d)$
        \STATE Update downstream priorities in $PQ$
    \ELSE
        \STATE $PQ.\textsc{Push}(d)$ with reduced priority
    \ENDIF
\ENDWHILE
\RETURN $S$
\end{algorithmic}
\end{algorithm}

\section{Experimental Details}

\subsection{Implementation Details and Hyperparameters}
\label{app:implementation_hparams}

We implement the environment generator, canonical parser, symbolic verifiers, repair operators, event-driven simulator, and PPO training pipeline in Python. 
Each synthetic environment is stored as a structured scaffold containing page templates, navigation edges, database schemas and records, marker specifications, task instructions, and programmatic completion constraints. 
All baselines use the same raw scaffolds and train the same compact DOM-grounded policy unless otherwise stated.

The policy contains fewer than 10M parameters and scores only DOM-grounded candidate actions extracted from the current rendered observation. 
We use Adam for optimization. 
Unless otherwise stated, PPO uses discount factor \(\xi=0.99\), GAE parameter \(\lambda_{\mathrm{GAE}}=0.95\), clipping threshold \(0.2\), value-loss coefficient \(0.5\), entropy coefficient \(0.01\), learning rate \(3\times 10^{-4}\), and gradient clipping threshold \(0.5\). 
Advantages are normalized within each PPO batch.

The dense reward coefficients in Eq.~\eqref{eq:reward} are fixed across domains and selected on the validation split: \(\alpha=0.5\), \(\gamma=0.2\), and \(\eta=0.01\). 
Here, \(\alpha\) controls the contribution of verified progress shaping, \(\gamma\) penalizes rejected marker-triggered state updates, and \(\eta\) is a small per-step cost that discourages unnecessarily long trajectories. 
The maximum episode horizon is \(H_{\max}=40\). 
Each PPO update uses \(4096\) rollout steps, minibatch size \(512\), and \(4\) optimization epochs. 
We train each policy for \(1.0\times 10^6\) environment steps unless otherwise stated.

For all reported policy experiments, we use the same train/validation/test environment split described in Section~\ref{sec:experiments}. 
Hyperparameters are selected using validation environments only and are then fixed for held-out synthetic evaluation and transfer evaluation.

\subsection{Compute Resources}
\label{app:compute_resources}

All experiments were run on a shared Linux compute cluster using CPU rollout workers and single-GPU training jobs. 
Each PPO training job used one NVIDIA A10G GPU with 24GB memory, 16 CPU cores, 64GB system RAM, and approximately 200GB local scratch storage for rollout logs and checkpoints. 
Synthetic environment generation and verification primarily used CPU workers plus LLM API calls, while PPO training used GPU workers for policy optimization and CPU workers for parallel environment rollout.

A single PPO run for one training condition takes approximately 3.2 hours on one NVIDIA A10G GPU, corresponding to about 3.2 GPU-hours. 
The main learning-curve experiments use five training conditions and three random seeds, for a total of approximately 48 GPU-hours. 
The component ablation experiments use seven ablated variants and three random seeds, requiring approximately 67 GPU-hours. 
The additional diagnostic experiments, including defect-impact analysis, dense-reward alignment, simulator tradeoff analysis, and failure attribution, require approximately 35 GPU-hours in total. 
Transfer evaluation on WebArena, WebShop, and MiniWoB++ requires approximately 8 GPU-hours and 90 CPU-hours because the learned policy does not call an LLM at evaluation time.

Environment construction uses 500 raw synthetic environments and invokes LLM-based verifiers only during offline scaffold verification and repair. 
For our full verification pipeline, the average curation time is 18 minutes per environment, as reported in Table~\ref{tab:rq1_env_quality}. 
This corresponds to approximately 150 CPU-hours for the full synthetic environment suite, excluding parallelization overhead. 
The event-driven simulator cost and token usage are reported in Figure~\ref{fig:rq4_simulation_tradeoff}. 
Across all reported environment-construction runs, we used approximately 18 million LLM input/output tokens for scaffold generation, verification, and repair.

\section{Additional Experimental Results}
\label{app_exp}
\subsection{Does the learned policy transfer beyond synthetic environments?}
\label{app:transfer}

The main experiments evaluate whether verification improves learning on held-out
synthetic environments. We additionally test whether the learned policy transfers
to external web-agent benchmarks under a unified DOM-grounded action interface.
The goal of this experiment is not to claim that a small policy universally
dominates frontier LLM agents under their native multimodal browser interfaces,
but to test whether verified synthetic training produces reusable interaction
skills when all methods are evaluated under the same observation and action
protocol.

\paragraph{Protocol.}
We evaluate on three external benchmarks: WebArena-compatible tasks, WebShop,
and MiniWoB++. For WebArena, we use a DOM-compatible subset that does not require
private credentials, file uploads, or visual-only information unavailable to the
compact policy. Task goals are not rewritten, and benchmark success criteria are
preserved. All evaluated methods receive the same textual task instruction and
the same serialized DOM observation. All methods select from the same
DOM-grounded action set, including click, type, select, and navigation actions.
The compact policies do not call an LLM during evaluation. GPT-4 baselines use
the same action interface and are evaluated with a fixed action budget.

\begin{table}[t]
\centering
\small
\setlength{\tabcolsep}{4.5pt}
\begin{tabular}{lccc}
\toprule
\textbf{Benchmark} & \textbf{\# Tasks} & \textbf{Observation} & \textbf{Success Criterion} \\
\midrule
WebArena-compatible & 180 & Serialized DOM + task & Original programmatic evaluator \\
WebShop & 500 & Product-page DOM + task & Original purchase-match evaluator \\
MiniWoB++ & 560 & DOM + task & Original environment reward \\
\bottomrule
\end{tabular}
\caption{
Transfer-evaluation protocol. We use a unified DOM-grounded interface for all
methods. WebArena results are reported on the DOM-compatible subset described in
the text; WebShop and MiniWoB++ use their original success evaluators.
}
\label{tab:transfer_protocol}
\end{table}

\paragraph{Baselines.}
We compare against GPT-4 direct prompting, GPT-4 with ReAct-style prompting, a
small policy trained by behavior cloning on synthetic trajectories, PPO trained
on raw synthetic environments, PPO trained on verified environments with terminal
rewards only, and our full method. GPT-4 baselines use temperature $0$, a maximum
of 20 actions per episode for WebArena-compatible tasks, and the same DOM action
schema as the compact policy. For learned policies, no external benchmark
fine-tuning is performed.

\begin{table*}[t]
\centering
\small
\resizebox{\textwidth}{!}{%
\begin{tabular}{lccccc}
\toprule
\textbf{Method}
& \textbf{WebArena SR$\uparrow$}
& \textbf{WebShop SR$\uparrow$}
& \textbf{MiniWoB++ SR$\uparrow$}
& \textbf{Eval LLM Calls$\downarrow$}
& \textbf{WebArena Step$\downarrow$} \\
\midrule
GPT-4 Direct Prompting
& $10.6{\pm}2.1$
& $32.5{\pm}2.7$
& $41.2{\pm}2.9$
& $17.5$
& $17.5{\pm}0.6$ \\
GPT-4 ReAct
& $15.3{\pm}2.6$
& $38.7{\pm}2.9$
& $47.6{\pm}3.1$
& $21.8$
& $14.3{\pm}0.5$ \\
Small Policy, Synthetic BC
& $11.9{\pm}2.3$
& $30.4{\pm}2.5$
& $39.8{\pm}2.8$
& $0.0$
& $16.8{\pm}0.7$ \\
PPO on Raw Synthetic Env.
& $12.4{\pm}2.4$
& $29.6{\pm}2.5$
& $38.9{\pm}2.8$
& $0.0$
& $17.2{\pm}0.6$ \\
PPO on Verified Env. + Terminal
& $14.7{\pm}2.5$
& $37.2{\pm}2.8$
& $48.5{\pm}3.1$
& $0.0$
& $16.0{\pm}0.5$ \\
\textbf{Ours}
& $\mathbf{18.6{\pm}2.8}$
& $\mathbf{43.8{\pm}3.0}$
& $\mathbf{55.4{\pm}3.1}$
& $\mathbf{0.0}$
& $\mathbf{15.1{\pm}0.5}$ \\
\bottomrule
\end{tabular}%
}
\caption{
Transfer beyond synthetic environments. All methods are evaluated under the same
DOM-grounded observation and action interface. The compact policy trained in
verified environments transfers better than policies trained on raw synthetic
environments, while requiring no LLM calls at evaluation time. The comparison to
GPT-4 baselines should be interpreted under this constrained DOM-only interface,
not as a claim of general superiority under native multimodal browser use.
}
\label{tab:transfer_results}
\end{table*}

\paragraph{Analysis.}
Verified synthetic training improves transfer on all three external benchmarks.
Compared with PPO trained on raw synthetic environments, our full method improves
success by $6.2$ points on WebArena-compatible tasks, $14.2$ points on WebShop,
and $16.5$ points on MiniWoB++. The largest gains appear on WebShop and
MiniWoB++, where the external environments share more structural similarity with
the synthetic training tasks. On WebArena-compatible tasks, the gains are smaller
but still consistent, suggesting that verification improves general interaction
skills rather than only fitting synthetic layouts. Importantly, the compact policy
uses zero LLM calls at evaluation time. We therefore interpret these results as
evidence that verified synthetic environments provide reusable supervision for
efficient policies under a common DOM-grounded interface.

\subsection{Does event-driven simulation reduce cost while preserving state fidelity?}
\label{sec:rq4_simulation}

A central design choice of our framework is to execute ordinary interface transitions deterministically and invoke constrained state writing only at sparse marker-triggered events.
To understand whether this design improves the cost--fidelity tradeoff, we compare our simulator against three alternatives:
(\textit{i}) a \textbf{Step-wise LLM} simulator that queries an LLM for every transition,
(\textit{ii}) a \textbf{Deterministic-only} simulator that never performs backend writes, and
(\textit{iii}) an \textbf{Unconstrained Marker LLM} simulator that invokes an LLM only at markers but does not validate the proposed state deltas against marker schemas and invariants.

\begin{figure*}[t]
    \centering
    \includegraphics[width=\textwidth]{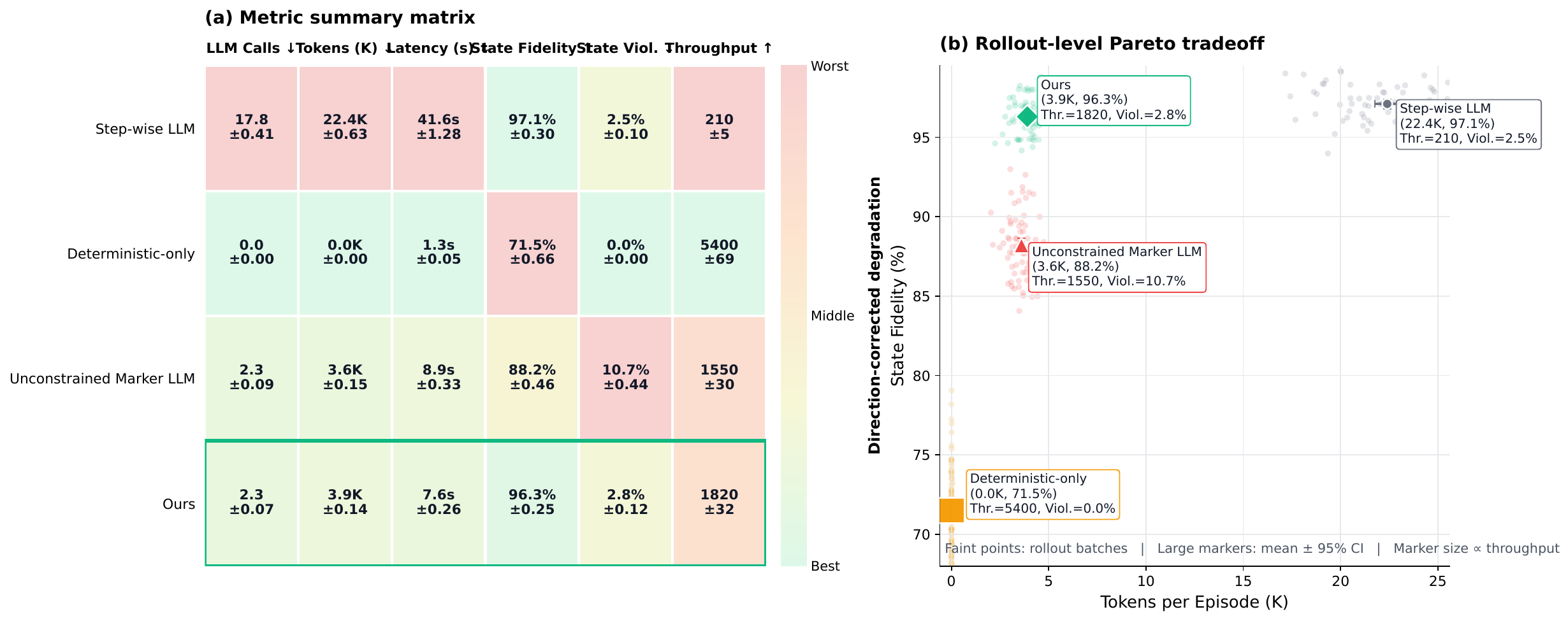}
    \caption{
    \textbf{Cost--fidelity tradeoff of simulators.}
    \textbf{Left:} metric summary matrix.
    Each cell reports the mean and 95\% confidence interval over rollout batches; color indicates metric-wise degradation after accounting for whether higher or lower is better.
    \textbf{Right:} rollout-level Pareto plot of token cost and state fidelity.
    Faint points denote individual rollout-batch observations; large markers denote simulator means with 95\% confidence intervals, and marker size is proportional to rollout throughput.
    Our event-driven simulator lies near the practical Pareto frontier: it preserves fidelity close to step-wise LLM simulation while using far fewer tokens and achieving much higher throughput.
    }
    \label{fig:rq4_simulation_tradeoff}
\end{figure*}

Instead of reporting a separate table and a separate Pareto plot, Fig.~\ref{fig:rq4_simulation_tradeoff} summarizes the full comparison in one figure.
The left panel reports the mean and 95\% confidence interval for six metrics:
average LLM calls per episode (\textbf{LLM Calls}),
token usage per episode (\textbf{Tokens}),
end-to-end episode latency (\textbf{Latency}),
state fidelity to the reference execution (\textbf{State Fidelity}),
state-invariant violation rate after writes (\textbf{State Viol.}),
and rollout throughput (\textbf{Throughput}).
The right panel visualizes the rollout-level Pareto tradeoff between token cost and state fidelity, where faint points show individual rollout-batch observations and large markers show mean performance with 95\% confidence intervals.

Figure~\ref{fig:rq4_simulation_tradeoff} shows that the step-wise LLM simulator achieves the highest state fidelity (\(97.1\%\)), but at a prohibitive cost: it requires \(17.8\) LLM calls and \(22.4\)K tokens per episode, resulting in the highest latency and the lowest rollout throughput.
At the other extreme, the deterministic-only simulator is extremely cheap and fast, but its fidelity drops to \(71.5\%\), indicating that removing backend writes entirely makes the environment too unrealistic for stateful web interaction.

The unconstrained marker simulator improves efficiency by reducing the number of LLM calls to \(2.3\) and the token cost to \(3.6\)K, but its lack of validation leads to a sharp increase in state violations (\(10.7\%\)) and substantially lower fidelity (\(88.2\%\)).
Our event-driven simulator achieves a better balance.
Compared with step-wise LLM simulation, it reduces token usage by \(82.6\%\) (from \(22.4\)K to \(3.9\)K) while preserving nearly the same fidelity (\(96.3\%\) vs.\ \(97.1\%\)), and improves rollout throughput by about \(8.7\times\) (from \(210\) to \(1820\) episodes per hour).
Compared with the unconstrained marker simulator, our method retains a similar call budget but improves fidelity by \(8.1\) points and reduces state violations by more than \(3.8\times\).

These results suggest that sparse marker triggering alone is not sufficient.
The key benefit comes from combining \emph{event-driven invocation} with \emph{state-safe validation}:
the former removes unnecessary per-step generation cost, while the latter prevents low-cost simulation from drifting away from faithful backend dynamics.
This is precisely the regime needed for scalable PPO training in synthetic web environments.

\subsection{Which defects most harm policy learning?}
\label{sec:defect_policy_impact}

The previous results show that verification improves environment executability and downstream PPO success.
We next ask a more diagnostic question: \emph{which types of scaffold defects are most harmful for policy learning?}
This analysis is important because defect frequency alone may be misleading.
A common semantic defect may be visually noticeable but harmless for learning, whereas a rare feasibility defect can corrupt the reward signal by making a task impossible under any policy.

\paragraph{Setup.}
Starting from verified environments, we construct controlled defect-mixture variants by reintroducing localized defects into pages, database bindings, markers, and task workflows.
For each variant, we train the same compact PPO policy under identical hyperparameters and measure the drop in held-out success rate relative to the fully verified environment:
\[
\Delta \mathrm{SR}(d)
=
\mathrm{SR}_{\mathrm{verified}}
-
\mathrm{SR}_{d},
\]
where \(d\) denotes a defect subtype.
For each subtype, we also measure its occurrence frequency, task-blocking rate, affected task fraction, and marker-write rejection rate.
This produces a defect-level impact profile that connects environment errors to downstream learning degradation.

\begin{figure*}[t]
    \centering
    \includegraphics[width=\textwidth]{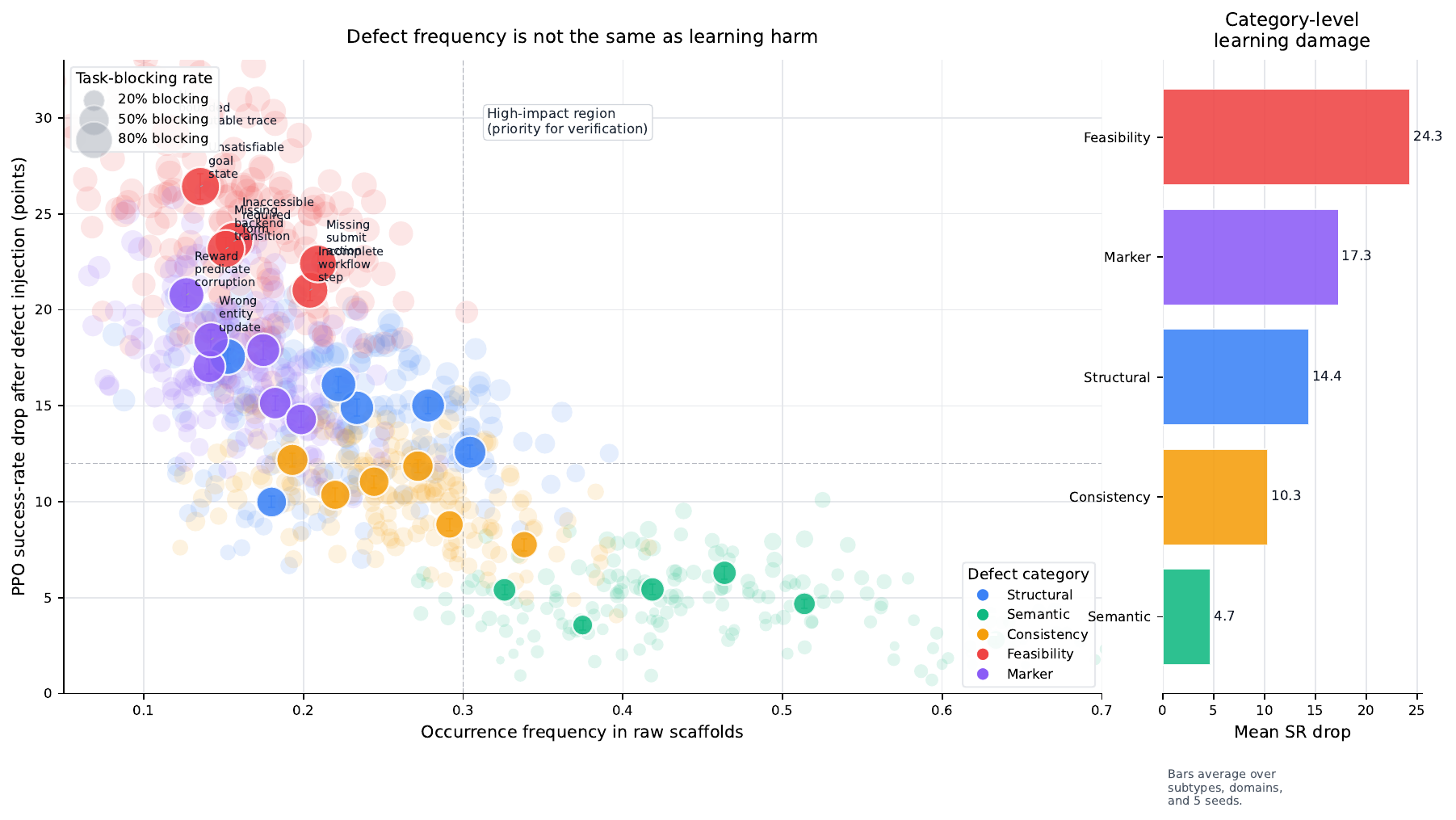}
    \caption{
    \textbf{Defect impact on policy learning.}
    Each point denotes one defect subtype measured across domains and random seeds.
    The \(x\)-axis reports how often the defect occurs in raw scaffolds, while the \(y\)-axis reports the PPO success-rate drop after injecting that defect into otherwise verified environments.
    Marker size indicates task-blocking rate and color denotes defect category.
    Feasibility and structural defects occupy the upper-impact region despite being less frequent than semantic defects.
    This shows that the most damaging defects are those that invalidate executable workflows or corrupt backend-grounded progress, rather than those that merely affect surface plausibility.
    }
    \label{fig:defect_policy_impact}
\end{figure*}

Figure~\ref{fig:defect_policy_impact} reveals a clear mismatch between defect frequency and learning harm.
Semantic defects are the most frequent in raw scaffolds, but many of them have modest impact on PPO because they do not necessarily prevent state progress.
In contrast, feasibility defects such as missing submit actions, inaccessible forms, and unsatisfiable completion constraints produce the largest success-rate drops.
Structural defects are also highly damaging when they disconnect required pages or remove navigation paths.
Marker defects occupy a second high-risk regime: they may not block navigation, but they corrupt backend updates and therefore make dense progress rewards unreliable.

These results provide a more fine-grained explanation for the ablation trends in Section~\ref{sec:rq3_components}.
The feasibility verifier is critical not because feasibility defects are the most common, but because each such defect creates a large amount of misleading negative experience for PPO.
Similarly, marker validation matters because even sparse state-write errors can poison the reward signal.
Therefore, verification should not be optimized only for reducing the total number of defects; it should prioritize defects with high task-blocking and reward-corrupting effects.

% =========================
% Experiment subsection
% =========================
\subsection{Is dense reward aligned with terminal success?}
\label{sec:dense_reward_alignment}

Dense rewards accelerate PPO training, but they are only useful if intermediate progress is aligned with final task completion.
A poorly designed reward may encourage local progress without completing the user instruction, leading to reward hacking.
We therefore evaluate whether our state-grounded dense reward is statistically calibrated with terminal success.

\paragraph{Setup.}
For each held-out task, we collect rollouts from partially trained PPO checkpoints and record the final progress potential \(\Psi_t(s_H)\), cumulative dense reward \(R_{\mathrm{dense}}\), and terminal success \(C_t(s_H)\).
We compare three reward signals:
\textsc{Surface-Heuristic}, which rewards visible UI changes such as clicks and form edits;
\textsc{LLM-Judge}, which asks an LLM to score partial task completion from rendered observations;
and \textsc{State-Grounded}, our reward computed from verified backend predicates.
For each reward, we measure calibration between predicted progress and terminal success, the area under the ROC curve (AUROC), Spearman correlation, expected calibration error (ECE), and the high-progress failure rate:
\[
\mathrm{HPF}
=
\Pr\bigl(C_t(s_H)=0 \mid \Psi_t(s_H) > 0.8\bigr).
\]
A well-aligned reward should have high AUROC and Spearman correlation, low ECE, and low high-progress failure rate.

\begin{figure*}[t]
    \centering
    \includegraphics[width=\textwidth]{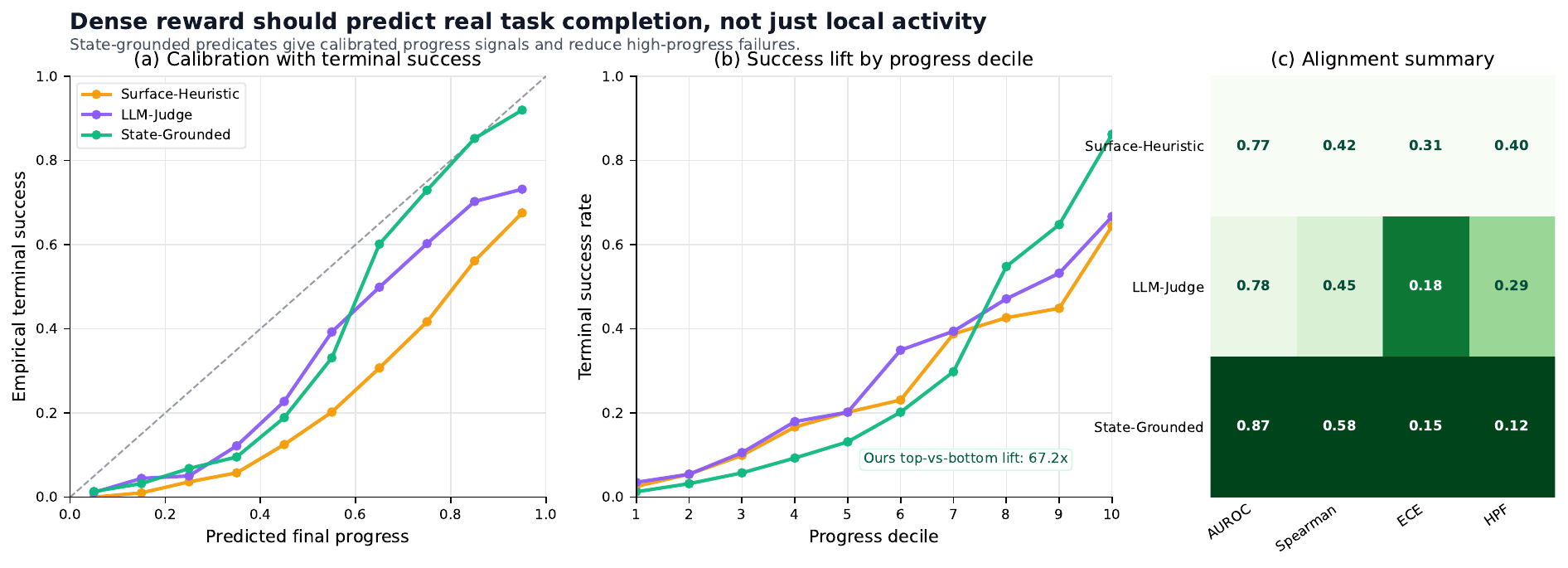}
    \caption{
    \textbf{Alignment between dense reward and terminal success.}
    \textbf{Left:} calibration curves between final progress score and empirical terminal success.
    A well-aligned dense reward should lie close to the diagonal.
    \textbf{Middle:} success rate by progress decile.
    Our state-grounded progress produces a monotonic success-lift pattern, whereas surface and LLM-based rewards assign high progress to many unsuccessful rollouts.
    \textbf{Right:} reward-alignment summary across domains, including AUROC, Spearman correlation, expected calibration error, and high-progress failure rate.
    State-grounded dense reward is both more predictive and better calibrated with terminal success, indicating that PPO receives intermediate supervision consistent with the true task objective.
    }
    \label{fig:dense_reward_alignment}
\end{figure*}

Figure~\ref{fig:dense_reward_alignment} shows that state-grounded dense reward is substantially better aligned with terminal success than surface-level or LLM-judged alternatives.
The calibration curve of our reward stays close to the diagonal, suggesting that a progress score of \(0.7\), for example, corresponds to an approximately \(70\%\) empirical chance of success.
In contrast, \textsc{Surface-Heuristic} is over-optimistic: many trajectories receive high reward for clicking, typing, or navigating, even when backend constraints remain unsatisfied.
\textsc{LLM-Judge} improves over surface heuristics but still overestimates progress in states where the rendered page looks plausible while hidden database fields are incorrect.

The decile analysis further confirms this trend.
For our state-grounded reward, terminal success increases almost monotonically with progress deciles, and the top decile has a much higher success rate than the bottom decile.
The high-progress failure rate is also much lower for our method, showing that the reward is less vulnerable to reward hacking.
These results support the central design choice of compiling task constraints into backend-state predicates: dense supervision should reward verified state progress, not merely plausible-looking interaction behavior.

\subsection{What are the failure modes of learned policies?}
\label{sec:policy_failure_attribution}

Success rate alone does not reveal whether a failed rollout is caused by the policy or by an invalid training environment.
This distinction is crucial for synthetic web-agent training.
If an episode fails because the scaffold is broken, the task is infeasible, or the backend update is inconsistent, PPO receives misleading negative feedback.
In contrast, if failure is caused by wrong navigation, wrong DOM grounding, or insufficient exploration, the failure is attributable to the learned policy and can be improved through training.

\paragraph{Attribution protocol.}
For each failed rollout, we assign one primary failure label using a deterministic diagnostic order.
First, we check whether the task admits a bounded executable trace under the current scaffold; if not, the failure is labeled as \textsc{Environment Invalidity}.
Second, if the rollout triggers a marker-write rejection or produces a backend invariant violation, it is labeled as \textsc{State-Update Violation}.
Third, if the rollout reaches high dense progress but fails the terminal constraint, it is labeled as \textsc{Reward Mismatch}.
Remaining failures are attributed to the policy: \textsc{Grounding Error} when the policy selects a wrong DOM element or fills a wrong field, \textsc{Planning Error} when it visits valid pages in an invalid order or misses a required subgoal, and \textsc{Timeout / Exploration} when the rollout does not make sufficient progress before the horizon limit.

Formally, for a failed trajectory \(\tau=(s_0,a_0,\ldots,s_H)\), we assign
\[
\mathrm{Attr}(\tau)
=
\begin{cases}
\textsc{Environment Invalidity}, 
& \nexists \pi_{1:H'} \ \text{s.t.}\ C_t(s_{H'})=1,\\
\textsc{State-Update Violation},
& \sum_k \epsilon_k > 0 \ \text{or}\ \mathrm{Inv}(s_H)=0,\\
\textsc{Reward Mismatch},
& \Psi_t(s_H)>\tau_{\psi}\ \text{and}\ C_t(s_H)=0,\\
\textsc{Grounding Error},
& \exists k: a_k \in \mathcal A(o_k)\ \text{but targets an incorrect DOM element},\\
\textsc{Planning Error},
& \exists i: \phi_{t,i}(s_H)=0\ \text{for an unmet required subgoal},\\
\textsc{Timeout / Exploration},
& \text{otherwise}.
\end{cases}
\]
We use \(\tau_{\psi}=0.8\) in all experiments.

\begin{figure*}[t]
    \centering
    \includegraphics[width=\textwidth]{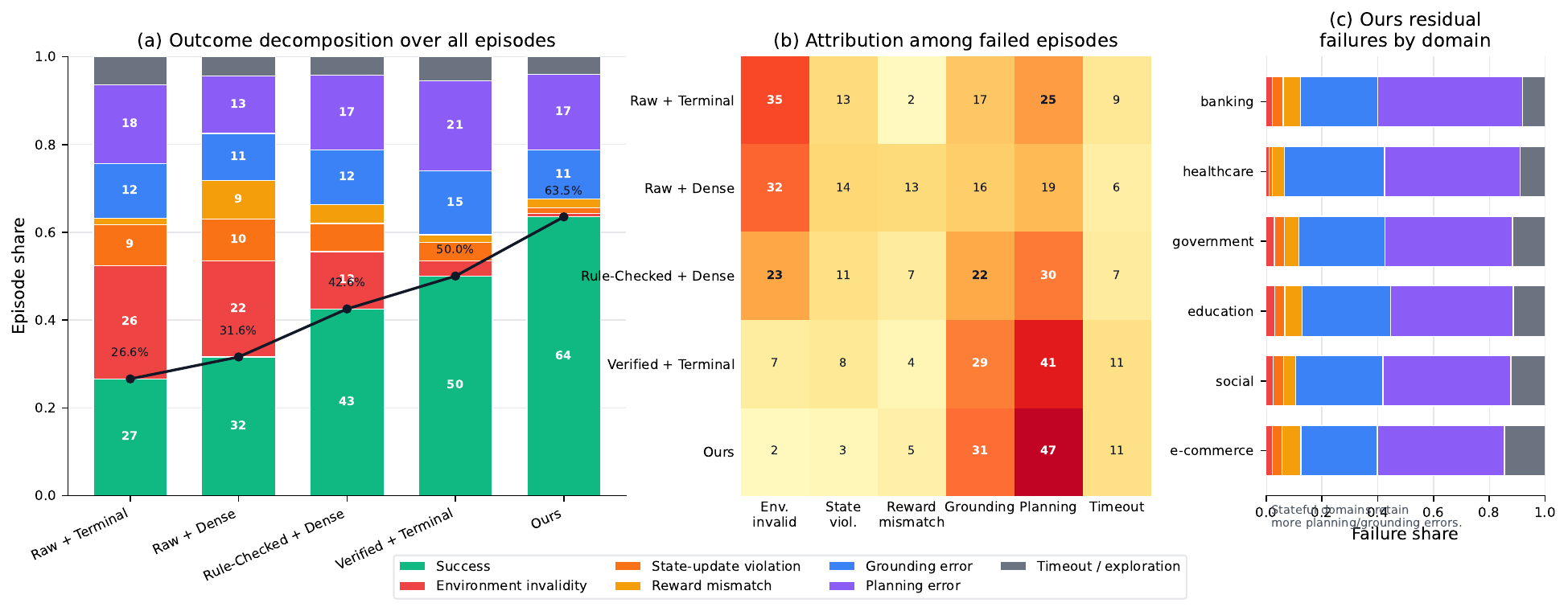}
    \caption{
    \textbf{Policy failure attribution.}
    \textbf{Left:} outcome decomposition over all evaluated episodes.
    Raw synthetic environments contain many environment-induced failures, making PPO feedback noisy.
    Verification sharply reduces environment invalidity and state-update violations.
    \textbf{Middle:} failure attribution conditioned on failed episodes.
    After verification, most remaining failures become policy-attributable planning and grounding errors.
    \textbf{Right:} domain-level residual failures for our full method.
    Harder stateful domains such as banking and healthcare retain more planning and grounding failures, suggesting where future policy improvements are needed.
    }
    \label{fig:policy_failure_attribution}
\end{figure*}

\paragraph{Results.}
Figure~\ref{fig:policy_failure_attribution} shows that raw synthetic environments produce a large fraction of non-policy failures.
In particular, \textsc{Environment Invalidity} and \textsc{State-Update Violation} dominate the failure distribution when PPO is trained on raw scaffolds.
This indicates that the policy is often penalized for tasks that are not executable or for backend transitions that are not faithfully simulated.
Adding dense rewards to raw environments does not fully solve this problem; it can even introduce additional \textsc{Reward Mismatch} failures, because progress predicates are unreliable when the underlying scaffold is inconsistent.

After verification and marker validation, the failure profile changes substantially.
Environment-induced failures become rare, and most remaining failures are attributable to the policy itself.
The full method therefore produces a cleaner training signal: failed episodes mostly correspond to wrong navigation, wrong element grounding, or insufficient long-horizon planning.
This shift is important because these are learnable policy errors, whereas broken scaffolds and invalid backend updates are not.
The domain-level analysis further shows that banking, healthcare, and government tasks retain more residual failures, mainly because they contain longer workflows and more state-dependent constraints.
These results suggest that verified environments do not merely improve average success rate; they also make failure feedback more interpretable and more useful for policy optimization.

\subsection{Is executability improvement confirmed by an independent audit?}
\label{app:independent_audit}

The feasible-task metric in the main paper is computed by bounded trace analysis.
Because this analysis is also used during repair, a natural concern is that the
reported executability improvement may be partly induced by the verifier itself.
We therefore perform an independent audit using two sources of validation that
are not used by the repair loop.

\paragraph{Setup.}
We sample 120 held-out environments and 720 tasks after applying each
environment-construction method. We evaluate executability using: (i) the
internal feasibility verifier used in the main experiments, (ii) a held-out
bounded trace analyzer with a separately implemented action enumerator and a
different search ordering, and (iii) author-audited replay on a randomly sampled
subset of 240 tasks. The author audit is used only as a sanity check and is not
used for model selection, repair, or hyperparameter tuning. We also report the
false-feasible rate, defined as the percentage of tasks marked feasible by the
internal verifier but judged infeasible by both the held-out analyzer and the
author audit.

\begin{table*}[t]
\centering
\small
\resizebox{\textwidth}{!}{%
\begin{tabular}{lccccc}
\toprule
\textbf{Method}
& \textbf{Internal Feasible$\uparrow$}
& \textbf{Held-out Feasible$\uparrow$}
& \textbf{Author-Audited Exec.$\uparrow$}
& \textbf{False Feasible$\downarrow$}
& \textbf{Trace Agreement$\uparrow$} \\
\midrule
No Verification
& $48.6{\pm}1.4$
& $46.1{\pm}1.6$
& $44.8{\pm}2.4$
& $12.4{\pm}1.5$
& $83.2{\pm}2.1$ \\
Rule-Based
& $53.2{\pm}1.5$
& $51.4{\pm}1.7$
& $49.6{\pm}2.6$
& $10.7{\pm}1.4$
& $84.9{\pm}1.9$ \\
Single-LLM
& $70.4{\pm}1.3$
& $67.8{\pm}1.5$
& $65.9{\pm}2.2$
& $6.8{\pm}1.1$
& $89.4{\pm}1.6$ \\
Self-Consistency
& $79.6{\pm}1.1$
& $77.2{\pm}1.3$
& $75.4{\pm}2.0$
& $5.2{\pm}0.9$
& $91.1{\pm}1.4$ \\
AutoGen
& $88.7{\pm}0.9$
& $86.9{\pm}1.0$
& $84.8{\pm}1.9$
& $3.9{\pm}0.7$
& $93.8{\pm}1.1$ \\
\textbf{Ours}
& $\mathbf{94.8{\pm}0.7}$
& $\mathbf{92.9{\pm}0.9}$
& $\mathbf{91.7{\pm}1.8}$
& $\mathbf{2.3{\pm}0.6}$
& $\mathbf{96.4{\pm}0.8}$ \\
\bottomrule
\end{tabular}%
}
\caption{
Independent executability audit. The feasible-task improvement is confirmed by a
held-out trace analyzer and author-audited replay. The low false-feasible rate
indicates that our main feasible-task metric is not merely an artifact of the
repair-time verifier.
}
\label{tab:independent_audit}
\end{table*}

\paragraph{Analysis.}
The held-out analyzer and author audit preserve the same method ranking as the
internal metric. Our method achieves $92.9\%$ held-out feasible tasks and
$91.7\%$ author-audited executability, only slightly below the internal
$94.8\%$ estimate. The false-feasible rate also drops from $12.4\%$ for raw
environments to $2.3\%$ for our method. This suggests that the repair loop does
not merely optimize for the internal feasibility checker; it produces workflows
that remain executable under an independently implemented analyzer and manual
replay.

\subsection{How reliable are the verifiers and repair operators?}
\label{app:verifier_reliability}

The verification stage is useful only if detected defects correspond to real
environment errors and repairs do not introduce new inconsistencies. We therefore
evaluate verifier reliability and repair reliability on an audited defect set.

\paragraph{Setup.}
We sample 80 raw environments and construct an audited defect set by manually
checking page graphs, database bindings, markers, and task workflows. The audit
contains 1,286 labeled defects across symbolic, structural, semantic,
consistency, feasibility, and marker categories. We evaluate each verification
method by precision, recall, and F1 against this audited set. For accepted
defects, we additionally measure repair success rate, defined as the fraction of
repairs that remove the target defect without creating another critical defect,
and false repair rate, defined as the fraction of repairs applied to audit-negative
defect reports.

\begin{table*}[t]
\centering
\small
\setlength{\tabcolsep}{5.0pt}
\begin{tabular}{lccccc}
\toprule
\textbf{Verifier}
& \textbf{Precision$\uparrow$}
& \textbf{Recall$\uparrow$}
& \textbf{F1$\uparrow$}
& \textbf{Repair Success$\uparrow$}
& \textbf{False Repair$\downarrow$} \\
\midrule
Rule-Based
& $96.1{\pm}1.2$
& $39.4{\pm}2.1$
& $56.0{\pm}2.0$
& $71.8{\pm}2.8$
& $1.2{\pm}0.4$ \\
Single-LLM
& $78.4{\pm}2.0$
& $66.7{\pm}2.4$
& $72.1{\pm}2.2$
& $74.3{\pm}2.6$
& $8.9{\pm}1.1$ \\
Self-Consistency
& $83.1{\pm}1.8$
& $73.6{\pm}2.2$
& $78.1{\pm}2.0$
& $80.2{\pm}2.4$
& $6.1{\pm}0.9$ \\
AutoGen
& $86.7{\pm}1.6$
& $79.5{\pm}2.0$
& $82.9{\pm}1.8$
& $84.6{\pm}2.1$
& $4.8{\pm}0.8$ \\
\textbf{Ours}
& $\mathbf{91.5{\pm}1.4}$
& $\mathbf{87.2{\pm}1.7}$
& $\mathbf{89.3{\pm}1.5}$
& $\mathbf{90.8{\pm}1.8}$
& $\mathbf{2.7{\pm}0.6}$ \\
\bottomrule
\end{tabular}
\caption{
Verifier and repair reliability on an audited defect set. Rule-based checking is
high precision but low recall. Single-pass LLM verification detects more defects
but also produces more false repairs. Our coordinated verifier improves recall
while keeping precision high and false repairs low.
}
\label{tab:verifier_reliability}
\end{table*}

\begin{table}[t]
\centering
\small
\setlength{\tabcolsep}{4.5pt}
\begin{tabular}{lcccc}
\toprule
\textbf{Category}
& \textbf{Precision$\uparrow$}
& \textbf{Recall$\uparrow$}
& \textbf{F1$\uparrow$}
& \textbf{Repair Success$\uparrow$} \\
\midrule
Symbolic
& $97.4$
& $93.1$
& $95.2$
& $96.5$ \\
Structural
& $91.2$
& $86.7$
& $88.9$
& $90.5$ \\
Semantic
& $88.6$
& $84.2$
& $86.3$
& $87.1$ \\
Consistency
& $90.4$
& $85.5$
& $87.9$
& $88.3$ \\
Feasibility
& $87.8$
& $91.6$
& $89.7$
& $92.1$ \\
Marker
& $92.3$
& $88.9$
& $90.6$
& $93.4$ \\
\bottomrule
\end{tabular}
\caption{
Category-level reliability of our verifier. Feasibility defects have slightly
lower precision but higher recall, which is desirable because missed feasibility
defects are especially harmful for policy learning.
}
\label{tab:category_reliability}
\end{table}

\paragraph{Analysis.}
The results explain why simple rule-based checking is insufficient: it rarely
hallucinates defects, but it misses many semantic, consistency, and feasibility
failures. Single-LLM verification improves recall but has a higher false repair
rate, which can introduce unnecessary edits to otherwise valid scaffold
components. Our method obtains the best F1 and repair success rate because
defect-triggered coordination asks only the relevant verifier to re-check
downstream effects, while priority-weighted repair scheduling avoids repairing
low-confidence isolated reports before high-impact dependencies are resolved.
The category-level results also support the design emphasis on feasibility and
marker validation: these categories are detected with high recall and repaired
with high success, reducing the chance that PPO receives misleading feedback
from impossible tasks or invalid backend updates.

\subsection{Are the gains due to a larger LLM verification budget?}
\label{app:llm_budget}

Our verification pipeline uses LLM-based semantic verifiers, so a natural
question is whether its advantage comes from better coordination or simply from a
larger LLM budget. We therefore compare verification methods under measured
LLM-call and token budgets. We also include an \textsc{Ours-Token-Matched}
variant that uses the same average token budget as the Single-LLM verifier by
running only one targeted coordination round and disabling optional low-severity
re-checks.

\begin{table*}[t]
\centering
\small
\resizebox{\textwidth}{!}{%
\begin{tabular}{lcccccc}
\toprule
\textbf{Method}
& \textbf{LLM Calls / Env.$\downarrow$}
& \textbf{Tokens / Env.$\downarrow$}
& \textbf{Time / Env.$\downarrow$}
& \textbf{Def.$\downarrow$}
& \textbf{Feasible$\uparrow$}
& \textbf{False Feasible$\downarrow$} \\
\midrule
Rule-Based
& $0.0$
& $0.0$K
& $0.5$m
& $9.8$
& $53.2$
& $10.7$ \\
Single-LLM
& $8.1$
& $41.8$K
& $12.0$m
& $6.2$
& $70.4$
& $6.8$ \\
Self-Consistency
& $40.0$
& $207.4$K
& $35.0$m
& $5.8$
& $79.6$
& $5.2$ \\
AutoGen
& $29.4$
& $154.2$K
& $28.0$m
& $5.1$
& $88.7$
& $3.9$ \\
Ours-Token-Matched
& $8.3$
& $42.5$K
& $11.5$m
& $4.7$
& $87.4$
& $3.4$ \\
\textbf{Ours}
& $22.6$
& $116.4$K
& $18.0$m
& $\mathbf{3.3}$
& $\mathbf{94.8}$
& $\mathbf{2.3}$ \\
\bottomrule
\end{tabular}%
}
\caption{
LLM-budget comparison for environment verification. Our full method uses fewer
tokens and less curation time than self-consistency and AutoGen while achieving
higher executability. Even when matched to the Single-LLM token budget, our
targeted coordination substantially improves feasible-task rate, suggesting that
the gain is not merely due to spending more LLM calls.
}
\label{tab:llm_budget}
\end{table*}

\paragraph{Analysis.}
Self-consistency spends the largest token budget because it repeatedly queries
independent verifiers, but it does not explicitly route downstream checks to the
defect categories most likely to be affected. AutoGen improves over
self-consistency but remains more expensive than our full method. In contrast,
our method uses targeted coordination and dependency-aware repair scheduling,
which reduces redundant re-checks. The \textsc{Ours-Token-Matched} variant is
particularly informative: with nearly the same token budget as Single-LLM, it
improves feasible-task rate from $70.4\%$ to $87.4\%$. This indicates that the
main benefit comes from structured verifier coordination and repair scheduling,
rather than from a larger LLM budget.

\subsection{Qualitative raw-to-verified case study}
\label{app:case_study}

We provide a representative example to illustrate how scaffold verification
changes the learning signal. The raw environment is an e-commerce website with
the task: ``Buy a wireless mouse under \$30 and place the order.'' The raw pages
look plausible, but the workflow is not executable because multiple scaffold
components disagree.

\begin{table*}[t]
\centering
\small
\setlength{\tabcolsep}{4.5pt}
\begin{tabular}{p{0.18\linewidth}p{0.25\linewidth}p{0.25\linewidth}p{0.20\linewidth}}
\toprule
\textbf{Defect}
& \textbf{Evidence in Raw Scaffold}
& \textbf{Repair}
& \textbf{Effect on Task} \\
\midrule
Broken navigation
& Product page links to \texttt{/cart}, but the declared navigation graph has no
reachable cart page.
& Add the missing cart route and reconnect it to the product and checkout pages.
& The bounded trace can reach the purchase workflow. \\
\midrule
Inconsistent price
& Product list shows the mouse as \$24.99, while the detail page binds the same
entity to \$34.99.
& Propagate the canonical database value \$24.99 to all rendered views.
& The price constraint in the task becomes evaluable and consistent. \\
\midrule
Missing backend transition
& The ``Add to cart'' button changes the visible page but does not update
\texttt{cart.items}.
& Attach an \texttt{add-to-cart} marker with typed arguments and write set
\texttt{\{cart.items\}}.
& Dense reward can credit verified cart insertion. \\
\midrule
Unsatisfiable completion constraint
& The completion predicate checks \texttt{order.status=placed}, but no checkout
button writes this field.
& Add a \texttt{place-order} marker and validate the state delta against order
schema invariants.
& Terminal success corresponds to an executable state transition. \\
\bottomrule
\end{tabular}
\caption{
Representative raw-to-verified repair example. The raw scaffold contains
multiple locally plausible but globally task-blocking defects. Verification and
repair convert the same task into an executable workflow with backend-grounded
progress predicates.
}
\label{tab:case_study}
\end{table*}

After repair, the shortest verified trace is:
\[
\texttt{Home}
\rightarrow
\texttt{SearchResults}
\rightarrow
\texttt{ProductDetail}
\rightarrow
\texttt{Cart}
\rightarrow
\texttt{Checkout}
\rightarrow
\texttt{Confirmation}.
\]
The corresponding progress predicates check whether the target product has been
visited, whether its canonical price satisfies the task constraint, whether the
correct entity has been inserted into \texttt{cart.items}, whether the checkout
form is valid, and whether the order state is updated to \texttt{placed}. In the
raw environment, PPO failures on this task are not attributable to the policy
because no policy can satisfy the completion constraint. In the verified
environment, failed rollouts are attributable to policy errors such as selecting
the wrong product, omitting checkout, or timing out before submitting the order.

\section{Limitations}
\label{app:limitations}

The event-driven simulator reduces cost by invoking constrained state writing only at marker-triggered operations. 
This is effective when persistent state updates are sparse, as observed in our diagnostic study, but may be less efficient for applications where nearly every interaction changes backend state. 

\section{Broader Impacts and Safeguards}
\label{app:broader_impacts_safeguards}

This work may have positive impacts by making web-agent training more reproducible, auditable, and less dependent on brittle live websites or expensive manual environment construction. 
Verified synthetic environments can help researchers distinguish policy failures from environment invalidity and can reduce unnecessary LLM calls during evaluation.

The same capability also has possible risks. 
More capable web agents could be misused for spam, unauthorized automation, credential abuse, synthetic phishing workflows, or other harmful web-scale actions if trained or deployed without safeguards. 

We can mitigate these risks in several ways. 
The proposed environments are sandboxed and use synthetic data rather than real user records. 
Persistent state changes are constrained by marker schemas, database invariants, and programmatic task constraints. 

\section{Declaration of LLM Usage}
\label{app:llm_usage}

LLMs are used to generate raw synthetic web scaffolds from domain-level site specifications, to support semantic verification of generated content and workflows, and to propose constrained state deltas for marker-triggered backend updates during synthetic training rollouts. 

In addition, LLMs are mainly used to help check for grammatical errors in writing.

\end{document}